\documentclass{article} %

\usepackage{iclr2027_conference,times}

\usepackage{amsmath,amsfonts,bm}

\def\eqref#1{equation~\ref{#1}}

\def\1{\bm{1}}

\DeclareMathAlphabet{\mathsfit}{\encodingdefault}{\sfdefault}{m}{sl}
\SetMathAlphabet{\mathsfit}{bold}{\encodingdefault}{\sfdefault}{bx}{n}

\usepackage{amsthm}
\newtheorem{theorem}{Theorem}[section]
\newtheorem{lemma}[theorem]{Lemma}

\newtheorem{proposition}[theorem]{Proposition}
\theoremstyle{definition}

\usepackage[dvipsnames]{xcolor}
\usepackage{bbm}
\usepackage{amssymb}

\usepackage{amsmath}

\usepackage{booktabs}

\usepackage{graphicx}

\usepackage{tikz}
\usetikzlibrary{positioning, calc}

\usepackage{subcaption} 

\usepackage{wrapfig}

\usepackage{enumitem}

\newif\ifarxiv
\arxivtrue

\usepackage{hyperref}
\usepackage{url}

\title{The Weakest Link: Distilling LLM Reasoning with Worst-Case Constrained Reinforcement Learning}

\newcommand{\iclrAffil}[1]{%
  \makebox[\dimexpr0.5\textwidth-2\tabcolsep\relax][l]{%
    \textnormal{#1}}%
}

\author{%
Matthieu Zimmer\thanks{Equal contribution.}\\
\iclrAffil{Huawei Noah's Ark Lab}
\And
Xiaotong Ji\footnotemark[1]\\
\iclrAffil{Huawei Noah's Ark Lab}
\AND
Tu Nguyen\\
\iclrAffil{Huawei Heisenberg Research Center}
\And
Haitham Bou-Ammar\\
\iclrAffil{Huawei Noah's Ark Lab}\\
\textnormal{UCL Centre for AI}
}

\ifarxiv
\iclrfinalcopy
\fi

\begin{document}

\maketitle

\begin{abstract}
Distilling the reasoning capabilities of large language models (LLMs) into smaller students is a central challenge for efficient deployment. Current approaches face a fundamental tension: optimizing purely for verifiable task rewards (e.g., via GRPO) leads to \textit{reward hacking}, where students arrive at correct final answers through flawed intermediate logic, while regularizing with soft divergence penalties against a teacher (e.g., KL-based distillation) dilutes task performance and, critically, allows the student to \textit{compensate} for severe logical violations at one step with high teacher agreement at others. We argue that this averaging is fundamentally misaligned with the nature of reasoning: a chain-of-thought is only as valid as its \textit{weakest link}. Motivated by this observation, we formulate reasoning distillation as a constrained reinforcement learning problem in which the task reward is maximized subject to a worst-case constraint on the teacher log-likelihood along every prefix of the trajectory. To avoid the prohibitive cost of dual Lagrangian solvers and the test-time teacher dependence of state-augmented methods such as Saute, we derive an unaugmented constrained MDP whose reward transformation preserves the hard-constraint semantics, admits a low-variance policy gradient decomposition into single-step and long-term terms, and provably satisfies the worst-case constraint almost surely in the penalty limit. Through extensive experiments on mathematical reasoning and code generation tasks, we demonstrate that our method significantly expands the accuracy-fidelity Pareto front. By matching the high Final Answer Correctness of pure RL and drastically reducing teacher constraint violations, we ultimately achieve the highest rigorous Reasoning Success Rate across all evaluated settings.
\end{abstract}

\section{Introduction}
\begin{figure}[ht]
    \centering
    \begin{tikzpicture}[
        boxa/.style={
            draw,
            rectangle,
            rounded corners,
            fill=teal!5,
            text width=0.47\linewidth,
            align=left,
            font=\tiny,
        },
        boxb/.style={
            draw,
            rectangle,
            rounded corners,
            fill=orange!8,
            text width=0.47\linewidth,
            align=left,
            font=\tiny,
        },
        box_hoz/.style={
            draw,
            rectangle,
            rounded corners,
            fill=violet!8,
            text width=0.47\linewidth,
            align=left,
            font=\tiny,
        },
        node distance=0.03cm
    ]
        
    \node (box2) [boxa] {
        \textbf{Ours}:\\[0.2cm]
        \setlength{\abovedisplayskip}{3pt}
        \setlength{\abovedisplayshortskip}{3pt}
        \setlength{\belowdisplayskip}{3pt}
        \setlength{\belowdisplayshortskip}{3pt}
        To determine what percentage of the whale's body length the combined length of the remoras is, we need to follow these steps:\par
        \textbf{1. Convert the length of the remoras from inches to feet:}
        \vspace{-0.05cm}
        {\setlength{\belowdisplayskip}{1pt} \[ 45 \text{ inches} \times \frac{1 \text{ foot}}{12 \text{ inches}} = 3.75 \text{ feet} \] }
        So, each remora is 3.75 feet long.\par
        \textbf{2. Calculate the total length of the remoras:}\vspace{-0.05cm}
        {\setlength{\belowdisplayskip}{1pt} \[4 \text{ remoras} \times 3.75 \text{ feet per remora} = 15 \]}
        \textbf{3. Determine the total length of the whale:}
        \vspace{-0.05cm}
        {\setlength{\belowdisplayskip}{1pt} \[
           300 \text{ feet}
        \]}
        \textbf{4. Calculate the percentage of the whale's body length that the combined length of the remoras represents:}\vspace{-0.05cm}
        \[
           \left( \frac{15 \text{ feet}}{300 \text{ feet}} \right) \times 100\%
        \]
        \textbf{5. Simplify the fraction:}\vspace{-0.25cm}
        \[
           \frac{15}{300} = 0.05
        \]\par
        \textbf{6. Convert the decimal to a percentage:}
        {\setlength{\belowdisplayskip}{0pt} \[
           0.05 \times 100\% = 5\%
        \]}
        Therefore, the combined length of the remoras is \(\boxed{5}\) percent of the whale's body length.
    };
    
    \node (box3) [boxb, right=of box2,yshift=0.582cm] {
        \textbf{GRPO (without teacher)}:\\[0.2cm]
        \setlength{\abovedisplayskip}{3pt}
        \setlength{\abovedisplayshortskip}{3pt}
        \setlength{\belowdisplayskip}{3pt}
        \setlength{\belowdisplayshortskip}{3pt}
Let's break down the problem step by step to find the percentage of the whale's body length that the combined length of the remoras represents.\par
First, we need to convert all units to the same unit. Let's use feet for this calculation.\par
The length of each remora is given as 45 inches. Since there are 12 inches in a foot, the length of each remora in feet is:\vspace{-0.05cm}
\[ \frac{45}{12} = 3.75 \text{ feet} \]
Since there are 4 remoras, the total length of the remoras is:\vspace{-0.05cm}
\[ 4 \times 3.75 = 15 \text{ feet} \]
Next, we add the length of the whale to the combined length of the remoras to get the total length of the whale:
\vspace{-0.05cm}
\[ 300 + 15 = 315 \text{ feet} \]
Now, we need to find what percentage the combined length of the remoras is of the whale's body length. This can be calculated using the formula for percentage:\vspace{-0.15cm}
\[ \left( \frac{15}{315} \right) \times 100 \]
Simplifying the fraction inside the percentage formula:\vspace{-0.05cm}
\[ \frac{15}{315} = \frac{1}{21} \]\vspace{-0.15cm}
\[ \left( \frac{1}{21} \right) \times 100 = \frac{100}{21} \approx 4.7619 \]
Rounding to the nearest whole number, we get: 
\( \boxed{5} \)
    };
    \coordinate (midpoint_2_3) at ($(box2.center)!0.5!(box3.center)$);

    \node (box1) [box_hoz, above=of midpoint_2_3,xshift=-3.42cm,yshift=2.955cm] {
        \textbf{Question}:\\
        Leilani saw a 300-foot whale with 4 45-inch remoras attached to it. What percentage of the whale's body length is the combined length of the remoras?
    };
    
    \end{tikzpicture}
    \caption{GRPO (right) reaches a wrong answer (4.7619) but applies a learned rounding trick to recover the correct answer (5).}
    \label{fig:qwen2_quali1_short}
\end{figure}

Large Language Models (LLMs) have achieved remarkable success in tasks requiring complex reasoning \citep{trinh2024solving,chervonyi2025goldmedalistperformancesolvingolympiad,guo2025deepseek}, but their sheer size makes them impractical for resource-constrained deployment. Distillation \citep{hinton2015distilling,zimmer:hal-01215273,xiong2024distillingmorphologyconditionedhypernetworksefficient} has traditionally been used to transfer this expertise to smaller models by minimizing output divergence (e.g., Kullback-Leibler divergence) between the student and the teacher \citep{minillm,distillm}. However, focusing solely on divergence forces students to blindly mimic complex reasoning paths that may exceed their representational capacity, while entirely ignoring the ultimate objective of the task. To close this gap, recent advancements reframe distillation as a reinforcement learning (RL) problem \citep{agarwal2024policy,grpo,huang2025treeopooffpolicymontecarlo}, where the student maximizes a verifiable task-specific reward (e.g., reaching the correct final math answer).

However, optimizing the task reward alone leads to a well-documented failure mode: \textbf{reward hacking}. The student discovers shortcuts that produce the correct \emph{final answer} through \emph{flawed intermediate reasoning}. Figure~\ref{fig:qwen2_quali1_short} illustrates this vividly: a GRPO-trained student computes a wrong numerical value (4.76\%) and then applies a learned ``rounding trick'' to recover the correct answer (5\%). The final answer is right, but the underlying logic is broken: it will not generalize.

The standard solution in the literature is to apply a soft regularization penalty \citep{agarwal2024policy}, balancing the RL reward with a step-wise divergence penalty against the teacher using a hyperparameter $\lambda$.
Unfortunately, this soft relaxation evaluates fidelity on \textit{average}. We argue that this is fundamentally misaligned with the nature of step-by-step reasoning. Because soft penalties compute an expected cost over the entire sequence, a student can easily learn to \textit{compensate} for a catastrophic logical error (which incurs a high penalty) by generating highly probable, boilerplate text elsewhere in the response. In logical reasoning, however, a chain-of-thought is only as strong as its \textbf{weakest link}. A single invalid step invalidates the entire path, regardless of how faithful the rest of the generation is to the teacher's stylistic manifold.

Motivated by this observation, we formulate reasoning distillation as a constrained Markov Decision Process (MDP) with a worst-case average constraint on the teacher log-likelihood. We require this constraint to be satisfied \textit{almost surely} during training.  
This acts as a strict guardrail: if the student generates any prefix that severely violates the teacher's manifold, the exploratory trajectory is immediately deemed infeasible and penalized. By transforming this threshold into a hard boundary rather than a soft preference, we prevent the optimizer from exploring and exploiting compensatory reasoning hacks.

Solving such constrained problems in the standard constrained-RL framework typically relies on dual Lagrangian methods \citep{10.5555/3305381.3305384,boyd2004convex,altman1999constrained}, which are often impractical at LLM scale due to the max--min inner-loop over the teacher. Alternatively, state-augmentation methods such as Saute \citep{sootla2022saute,sootla2022enhancing} avoid duality but require tracking a remaining budget variable at every step, which would force the student to query the teacher \emph{at test time}, therefore defeating the purpose of distillation. We address both limitations by leveraging the history-conditioned structure of LLM policies: the remaining budget is a deterministic function of the generated history and therefore need not be exposed as an augmented policy state. We further demonstrate that this unaugmented objective admits a low-variance policy gradient decomposition, splitting the update into a \emph{single-step} term (exploiting the token-level structure of the reward) and a \emph{long-term} credit-assignment term. This ensures that the student can be trained efficiently and deployed completely independently of the teacher while maintaining theoretical guarantees of constraint satisfaction in the penalty limit.

Our contributions are summarized as follows:
\begin{itemize}[leftmargin=1.5em, topsep=0pt, itemsep=0pt]
    \item We identify the "weakest link" flaw in soft-regularized distillation, where students mask severe logical errors by maximizing average sequence likelihood.
    \item We formulate reasoning distillation as a worst-case Constrained RL problem, introducing an unaugmented MDP that enforces trajectory-level hard constraints without requiring dual optimization or test-time teacher access.
    \item We detail a low-variance policy gradient decomposition that provides stable, intermediate token-level signals to guide the student along authorized reasoning paths.
    \item We demonstrate empirically that our method breaks the trade-off between final answer correctness and teacher compliance, achieving superior reasoning success rates across multiple LLM families on coding and mathematical benchmarks.
\end{itemize}

\section{Background}

\subsection{Distillation as Policy Optimization}
Knowledge distillation typically transfers capabilities from a vast teacher model $\mu$ to a compact student model $\pi$ by minimizing a divergence metric, often the Kullback-Leibler (KL) divergence, forcing the student to imitate the teacher's output distribution \citep{hinton2015distilling, minillm}. While this ensures fidelity, it ignores task-specific performance: a student might try to mimic a highly complex reasoning chain it lacks the capacity to complete, failing the fundamental task \citep{zhang-etal-2025-towards-law}. 

To remedy this, recent frameworks merge distillation with Reinforcement Learning (RL), recasting the problem as policy optimization. The student is trained to maximize an expected, verifiable task reward $R$ (e.g., matching the ground-truth final answer in a math problem) while being regularized by a sequence-level divergence $D(\pi,\mu)$ against the teacher:
\begin{equation}\label{eq:lag}
\max_{\pi} \mathbb{E}_\pi \left[ R - \lambda D(\pi, \mu) \right],
\end{equation}
where $\lambda>0$ acts as a soft penalty controlling the trade-off. 

\textbf{The Weakest Link problem:} In practice, \eqref{eq:lag} is fundamentally ill-suited for reasoning tasks. The divergence term $D(\pi,\mu)$ is typically evaluated as an expected sum over the trajectory. Because it is a \textit{soft average}, the optimization landscape permits \textbf{compensation}. A student model can undertake a catastrophic logical leap, incurring a large localized penalty, but offset this cost by generating long strings of highly probable, "safe" tokens that strictly adhere to the teacher's manifold. In step-by-step reasoning, this averaging effect allows reward hacking to survive: a chain of thought is only as correct as its weakest logical link. Resolving this requires a shift from soft expectations to hard, sequence-level constraints.

\subsection{Constrained Reinforcement Learning}
\label{subsec:constrained_MDP}
Constrained Reinforcement Learning (CRL) offers a mathematical framework for optimizing a primary reward while strictly forbidding certain behaviors \citep{10.5555/3305381.3305384,ji2025almost}. We formally define a Constrained MDP as a tuple $\mathcal{M}_d = \langle \mathcal{S}, \mathcal{A}, \mathcal{P}, R, C, \gamma, d \rangle$. In the context of LLM distillation, the state $\mathbf{s}_t$ corresponds to the prompt and the generated prefix, the action $\mathbf{a}_t$ is the next token generated by the student, $\mathcal{P}$ encodes the deterministic transition kernel mapping $(\mathbf{s}_t, \mathbf{a}_t) \rightarrow \mathbf{s}_{t+1}$, and $R$ is the task-specific reward (typically sparse, given at the end of the sequence). We define $C(\mathbf{s}_t)$ as a strictly positive per-step cost (e.g., the negative log-likelihood of the generation under the teacher $\mu$). 

Rather than softly regularizing the expected cost, a strict CRL formulation enforces that the accumulated cost remains below a predefined budget $d$ for every valid trajectory:
\begin{equation}\label{eq:constrained_mdp_saute}
\max_{\pi} \mathbb{E}_\pi \left[ \sum_{t=0}^\infty \gamma^t R(\mathbf{s}_t, \mathbf{a}_t) \right] 
\text{s.t.} \quad \forall T, \quad \sum_{t=0}^{T-1} C(\mathbf{s}_t) \leq d \text{\,\,\,\,\,\,\,\,\,\textbf{almost surely}}.
\end{equation}
Traditional methods for solving Eq.\ref{eq:constrained_mdp_saute} rely on Lagrangian duality, effectively dynamically tuning a time-dependent $\lambda$ parameter in Eq.\ref{eq:lag}. However, at LLM scale, maintaining and updating dual variables through inner-loop optimization is notoriously unstable and computationally prohibitive. 

An elegant alternative that avoids dual optimization is the method of state augmentation, known as Saute \citep{sootla2022saute}. Saute resolves the problem by formulating a new augmented MDP $\widetilde{\mathcal M}_{d}^{n} = \langle \tilde{\mathcal{S}}, \mathcal{A}, \tilde{\mathcal{P}}, \tilde{R}_{n}, \gamma, d \rangle$. It introduces an auxiliary tracking variable $\mathbf{z}_t$ that depletes the initial budget $d$ at each step ($\mathbf{z}_0 = d$, $\mathbf{z}_{t+1} = \mathbf{z}_t - C(\mathbf{s}_t)$), extending the state space to $\tilde{\mathcal{S}}=\mathcal{S} \times \mathcal{Z}$. The optimization then seamlessly proceeds over an unconstrained surrogate objective with a reshaped reward:
\begin{gather}\label{eq:saute_unconstrained}
\max_{\pi} \mathbb{E}_\pi \left[ \sum_{t=0}^\infty \gamma^t \tilde{R}_n(\mathbf{s}_t, \mathbf{z}_t, \mathbf{a}_t) \right], 
\tilde{R}_n(\mathbf{s}_T, \mathbf{z}_T, \mathbf{a}_T) = 
\begin{cases} 
R(\mathbf{s}_T, \mathbf{a}_T) & \text{if } \mathbf{z}_T \geq 0, \\
- n & \text{if } \mathbf{z}_T < 0.
\end{cases}
\end{gather}
where $n \gg R_{\max}$ provides a harsh, penalty for any trajectory that exhausts the budget. As $n \to \infty$, this penalty guarantees feasibility without saddle-point continuous optimization.

\textbf{Limitations for Distillation:} 
While Saute is an elegant approach, it is ill-suited for reasoning distillation for three reasons. First, its cumulative-cost constraint Eq.\ref{eq:constrained_mdp_saute} is poorly calibrated for sequence generation: because $C(\mathbf{s}_t) > 0$ at each step, longer trajectories accumulate larger budgets by construction, making the constraint length-dependent and biased against longer reasoning chains. Second, maintaining the augmented state $\mathbf{z}_t$ online requires evaluating $C$ at every step, which means querying the teacher \textit{at test time} and therefore fundamentally defeats the purpose of distillation. Third, its fixed penalty $-n$ for constraint violations provides no informative signal about \textit{which} token caused the violation, making credit assignment difficult. In the next section, we address all three issues by (i) reformulating the constraint as a worst-case trajectory-level bound, (ii) showing that the augmented state can be eliminated under history-conditioned policies, and (iii) deriving a low-variance policy gradient that exploits the token-level structure of the reward.

\section{Method}
To address the fundamentally sequential nature of reasoning, we formulate LLM distillation as a Constrained MDP. Unlike previous methods, our framework enforces a strict "weakest-link" boundary on teacher adherence, operates entirely without teacher access at test time, and introduces a theoretically grounded variance-reduction mechanism for stable optimization.

\subsection{The Weakest Link: Worst-Case Constrained RL}
\label{sec:worst_case_rl}

\begin{wrapfigure}[18]{r}{0.4\textwidth}
    \includegraphics[width=1.0\linewidth]{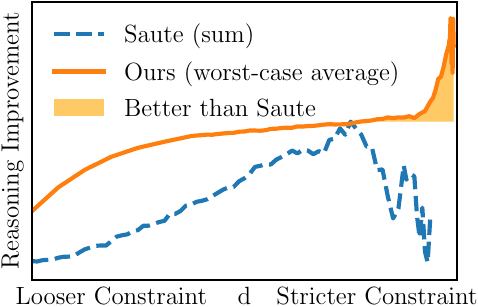}
    \caption{Reasoning Improvement across budget bounds. Constraining the worst-case average prevents local hallucination effectively, yielding strictly higher peak reasoning performance across optimal budget thresholds compared to a cumulative sum budget.}
    \label{fig:constraint_type}
\end{wrapfigure}

We begin by defining the per-step cost as the negative log-likelihood of the student's output under the teacher's distribution: $C(\mathbf{s}_t, \mathbf{a}_t) = - \log \mu(\mathbf{a}_t | \mathbf{s}_t)$.
We validate this proxy in Appendix~\ref{app:proxy_validation}, worst-case teacher likelihood correlates positively with judge-assessed reasoning validity in every run.
Standard CRL frameworks, such as Saute \citep{sootla2022saute}, constrain the cumulative sum of costs $\sum_t C_t \leq d$. However, because $C$ is strictly positive, a cumulative budget inherently penalizes longer trajectories, artificially discouraging the student from generating the extended chains of thought necessary for complex problem-solving. Furthermore, as discussed in Section 2, soft-regularization methods constrain the \textit{expected average} sequence cost, which permits reward-hacking via "compensation" (e.g., masking a severe logical error with highly probable surrounding text).

To prevent future compensation after a prefix has already crossed the boundary while remaining length-invariant, we propose optimizing the task reward subject to a \textbf{worst-case average constraint} over every prefix of the trajectory. Specifically, we require that at no point during the generation does the average cost exceed the threshold $d$:
\begin{equation}\label{eq:constrained_mdp_ours}
\max_{\pi} \mathbb{E}_\pi \left[ \sum_{t=0}^\infty \gamma^t R(\mathbf{s}_t, \mathbf{a}_t) \right] \quad \text{s.t.} \quad \max_{T \ge 1} \frac{1}{T} \sum_{t=0}^{T-1} C(\mathbf{s}_t, \mathbf{a}_t) \leq d \text{\,\,\,\,\,\,\,\,\,\textbf{almost surely}}.
\end{equation}

Empirically, restricting the worst-case trajectory effectively preserves reasoning integrity. As shown in Figure~\ref{fig:constraint_type}, maximizing reward under our worst-case average yields strictly higher reasoning improvement than optimal bounded cumulative sums. We formalize this advantage, and explicitly distinguish our approach from standard Lagrangian relaxations, in the following proposition:

\begin{proposition}[Anti-Compensation and Non-Equivalence]
Let $\tau$ be a trajectory of length $T$ with non-negative costs $C$.
\begin{enumerate}[label=(\alph*), leftmargin=1.5em, topsep=0pt, itemsep=0pt]
\item \textbf{Anti-Compensation:} If $\tau$ satisfies an expected average threshold $\frac{1}{T}\sum_{t=0}^{T-1} C_t \leq d$ (as induced by soft penalties), the maximum single-step error $C_{\max} = \max_t C_t$ can grow as $\mathcal{O}(T \cdot d)$, allowing unbounded local hallucinations. Conversely, under the worst-case average constraint $\max_{\tau \le T} \frac{1}{\tau} \sum_{t=0}^{\tau-1} C_t \le d$, the severity of a hallucination precisely at step $k$ is bounded by $(k +1) \cdot d$.
\item \textbf{Non-Equivalence:} There exists no constant scalar penalty $\lambda \geq 0$ such that the optimal policy of the unconditionally soft problem $\max_\pi \mathbb{E}_\pi[\sum \gamma^t (R - \lambda C)]$ matches the optimal policy of the worst-case constrained MDP in Eq.~\ref{eq:constrained_mdp_ours} for all $(R, C)$.
\end{enumerate}
\label{prop:anti_comp}
\end{proposition}
See Appendix \ref{app:anti_comp} for full proof. 

\subsection{Eliminating State Augmentation via Absorbing Infeasibility}

Although state-augmented methods like Saute can theoretically enforce trajectory constraints without inner-loop Lagrangian optimization, they rely on appending a remaining budget variable $\mathbf{z}_T$ to the MDP state. For distillation, tracking $\mathbf{z}_T$ online requires computing the teacher's likelihood $C(\mathbf{s}_t, \mathbf{a}_t)$ at every autoregressive decoding step. This mandates test-time access to the massive teacher model, completely neutralizing the computational benefits of distillation.

We solve this by directly modifying the reward function \emph{without} state augmentation. We construct a history-based MDP $\widehat{\mathcal{M}}^n_d = \langle \mathcal{S}, \mathcal{A}, \mathcal{P}, \hat{R_{n}}, \gamma, d \rangle$, where the constrained reward fuses the task reward with a feasibility penalty $n$ (when the constraint is broken, we enter an absorbing state):
\begin{equation}
\label{eq:our_prob}
\hat{R}_{n}(\mathbf{s}_T, \mathbf{a}_T) =
\begin{cases}
R(\mathbf{s}_T, \mathbf{a}_T) & \text{if } \max_{\tau \le T}\frac{1}{\tau}\sum_{t=0}^{\tau-1} C(\mathbf{s}_t, \mathbf{a}_t) \leq d, \\
- n & \text{otherwise.}
\end{cases}
\end{equation}

Why does this avoid the need for budget tracking? Because our worst-case constraint is inherently prefix-wise. We formalize this critical structural property:

\begin{lemma}[Absorbing Infeasibility]
Let the infeasible state space be defined as $\mathcal{S}_{\text{inf}} = \{ \mathbf{s}_T \mid \max_{\tau \le T} \frac{1}{\tau} \sum_{t=0}^{\tau-1} C > d \}$. For any generation step $\mathbf{a}_T \in \mathcal{A}$, the deterministic transition to the new prefix $\mathbf{s}_{T+1} = \mathbf{s}_T \oplus \mathbf{a}_T$ ensures that if $\mathbf{s}_T \in \mathcal{S}_{\text{inf}}$, then sequentially $\mathbf{s}_{T+1} \in \mathcal{S}_{\text{inf}}$.
\label{lemma:absorbing}
\end{lemma}

By Lemma~\ref{lemma:absorbing}, a single constraint violation irreversibly "kills" the trajectory. Because an LLM explicitly autoregressively conditions on the entire prefix $\mathbf{s}_T$, the control process is fully observable. The viability of the state is perfectly encoded within the history, eliminating the need to explicitly pass or compute an augmented state $\mathbf{z}_T$. Consequently, at test time, the student model operates entirely independently of the teacher. The teacher is merely queried \emph{a posteriori} during RL training rollouts to compute $\hat{R}_n$.
Under standard assumptions, this unaugmented objective preserves the core theoretical guarantees of constrained RL (Theorem \ref{thm:bellman}; proofs in Appendix \ref{app:bellman}):

\begin{theorem}
As the penalty $n \to \infty$, the optimal value function for $\widehat{\mathcal{M}}^n_d$ converges monotonically. Furthermore, if the limiting MDP admits an optimal policy with finite value, that policy satisfies the worst-case averaged constraint almost surely.
\label{thm:bellman}
\end{theorem}

\subsection{Policy Gradient Optimization and Variance Reduction} \label{subsec:pg_opt}

To optimize Eq.~\ref{eq:our_prob}, we maximize the expected discounted return $J_n(\theta) = \mathbb{E}_{\pi_\theta}[ \sum_{t=0}^\infty \gamma^t \hat{R}_{n}(\mathbf{s}_t, \mathbf{a}_t) ]$. Under the standard REINFORCE formulation \citep{sutton1999policy}, the gradient depends on the aggregate future return $R_t^n \triangleq \sum_{k=t}^\infty \gamma^{k-t}\hat{R}_{n}(\mathbf{s}_k,\mathbf{a}_k)$. However, this standard estimator suffers from high variance, especially in our constrained case, where a single token's choice to violate the boundary violently swings the entire future trajectory return to $-n$.

To stabilize optimization, we exploit the fact that evaluating our immediate penalized reward $\hat{R}_n(\mathbf{s}_t, \mathbf{a})$ does not require environment interaction, but just a forward pass of the frozen teacher over the current vocabulary. Using the recursion $R_t^n = \hat{R}_n(\mathbf{s}_t, \mathbf{a}_t) + \gamma R_{t+1}^n$, we elegantly decompose the gradient into a \emph{single-step} term evaluated precisely over the action space, and a \emph{long-term} Monte Carlo term (see Appendix \ref{app:derive} for the full derivation):
\begin{align}
\nabla_\theta J_n(\theta)
&=
\underbrace{
\mathbb{E}_{\pi_\theta}\!\Bigg[
\sum_{t=0}^\infty \gamma^t\,
\nabla_\theta
\Big(
\sum_{\mathbf{a}}
\pi_\theta(\mathbf{a}\mid \mathbf{s}_t)\,
\hat{R}_{n}(\mathbf{s}_t,\mathbf{a})
\Big)
\Bigg]
}_{(\nabla_\theta J_n)_{\textsc{single}}}
+
\underbrace{
\mathbb{E}_{\pi_\theta}\!\Bigg[
\sum_{t=0}^\infty \gamma^{t+1}
R_{t+1}^{n}
\,\nabla_\theta \log \pi_\theta(\mathbf{a}_t \mid \mathbf{s}_t)
\Bigg]
}_{(\nabla_\theta J_n)_{\textsc{long}}}.\notag
\end{align}

This decomposition provides intermediate token-level signals to guide the student carefully along authorized reasoning prefixes. Crucially, while this algebraic decomposition is universally valid, its practical variance-reduction benefits are uniquely impactful and tractable by our worst-case constraint formulation. Under standard trajectory-level constraints (such as global sums or expected averages), a sequence's feasibility is typically only determined at the terminal step. Consequently, the immediate reward at step $t$ contains no constraint information, and the massive variance of the constraint penalty remains entirely trapped in the stochastic long-term return $R_{t+1}^n$. 

In contrast, as established by Lemma~\ref{lemma:absorbing}, our prefix-wise constraint definitively triggers the $-n$ penalty the exact moment the boundary is crossed. By shifting the feasibility penalty into the immediate reward $\hat{R}_n(\mathbf{s}_t, \mathbf{a})$, we can analytically integrate it out over the vocabulary using a single forward pass of the teacher. Thus, this decomposition provides a mathematically guaranteed variance reduction for the immediate reward evaluation at every step (formalized in Appendix \ref{app:variance} in Theorem~\ref{thm:variance}).

Theorem~\ref{thm:variance} demonstrates that resolving the immediate step using the teacher distribution analytically removes the sampling variance associated with that specific token step. The long-term term $(\nabla_\theta J_n)_{\textsc{long}}$ is preserved to allow proper credit assignment, penalizing early tokens that steer the prefix toward an unavoidable constraint violation later in the trajectory.

\section{Experiments}
\label{sec:experiments}

We design our experiments to answer three core research questions:
\begin{itemize}[leftmargin=1.5em, topsep=0pt, itemsep=0pt]
    \item \textbf{RQ1 (The Trade-off):} Does enforcing a worst-case constraint prevent reward-hacking without sacrificing task performance?
    \item \textbf{RQ2 (Reasoning Quality):} Does eliminating compensatory reward-hacking lead to tangibly superior logical reasoning?
    \item \textbf{RQ3 (Anti-Compensation):} Can we empirically validate that soft baselines suffer from severe localized violations, and that our worst-case metric predicts reasoning failure?
\end{itemize}

\subsection{Experimental Setup}
\textbf{Tasks \& Models.} We evaluate on several mathematical reasoning datasets: MATH \citep{hendrycksmath2021}, Apple/GSM-Symbolic (main) \citep{mirzadeh2025gsmsymbolic} and GSM8K \citep{gsm8k}. We consider three distinct distillation settings to ensure generalization across model families and task difficulties: (1) \textit{Qwen2.5-1.5B} distilled from \textit{Qwen2.5-Math-7B-Instruct} on MATH; (2) \textit{Llama-3.2-3B} distilled from \textit{Llama-3.2-11B-Vision-Instruct} on MATH; and (3)
\textit{Qwen2.5-1.5B} distilled from \textit{Qwen2.5-Math-7B-Instruct} on GSM8K and evaluated on GSM-Symbolic.
We also performed experiments on code generation with different coding models that we report in Appendix~\ref{app:code_gen}.

\textbf{Baselines.} We compare our unaugmented constrained RL approach against: (1) \textbf{GKD} \citep{agarwal2024policy} and \textbf{Mini-LLM} \citep{minillm} for pure divergence minimization; (2) \textbf{GRPO} \citep{grpo} for pure task-reward maximization; and (3) \textbf{GKD-GRPO}, the standard soft Lagrangian relaxation \citep{agarwal2024policy} tuned across a wide spectrum of constraint penalties $\lambda \in \{0.001, 0.01, 0.1, 1.0, 10\}$.
For our method, we evaluate several worst-case average budget thresholds $d \in \{-\ln(1/4), -\ln(1/2), -\ln(3/4)\}$. For an exhaustive description of all baselines, prompt templates, training hyperparameters, and the exact evaluation protocols, please refer to Appendix \ref{sec:app_algo}.

\textbf{Metrics.} We report:
1) \textbf{Final Answer Correctness (FAC):} Verifies if the extracted \texttt{\textbackslash boxed\{\}} answer precisely matches the ground truth.
2) \textbf{Constraint Satisfaction (CS):} The percentage of test trajectories where the worst-case prefix average remains below the nominal threshold $d=-\ln(1/2)$. This measures fidelity.
3) \textbf{Reasoning Success Rate (RSR):} Beyond the final answer, we assess the absolute logical validity of the intermediate steps using an LLM-as-a-judge \citep{zheng2023judging}.
We also verified the judge labels on a small subset (see Appendix~\ref{app:judge_reliability}).

\begin{figure}[t]
    \centering
    \includegraphics[width=1.0\textwidth]{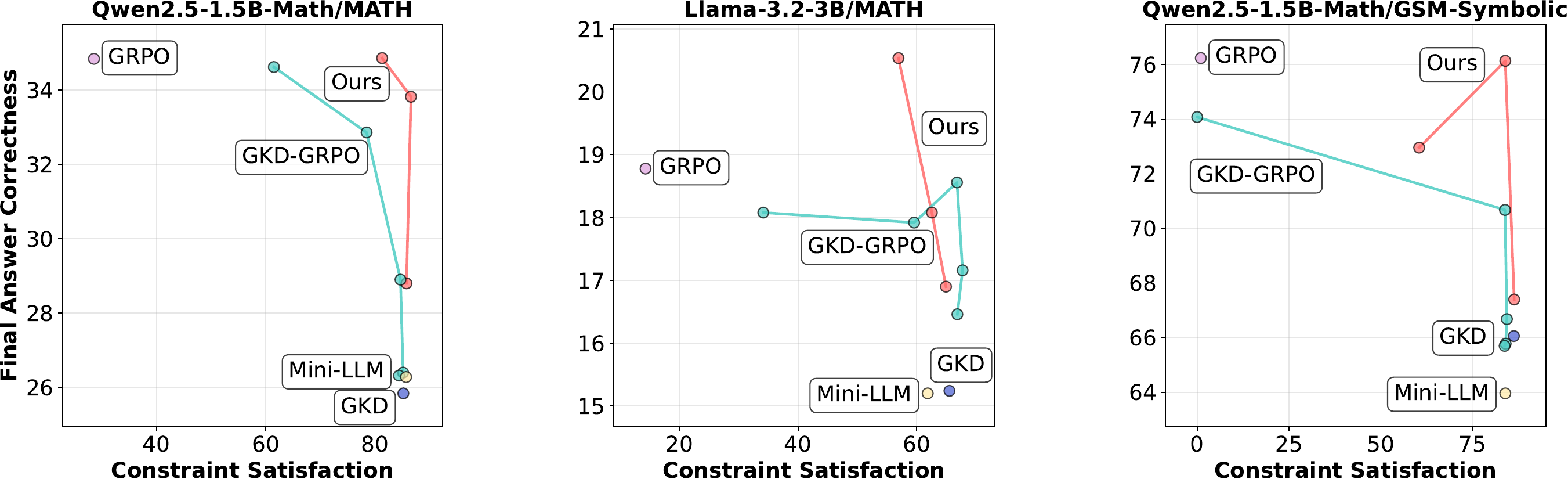}
    \caption{\textbf{Accuracy-Fidelity Pareto Front on the 3 distillation settings.} X-axis: Constraint Satisfaction (higher is better). Y-axis: Final Answer Correctness (higher is better). Soft-regularized baselines (GKD-GRPO line) force a harsh trade-off between fidelity and accuracy. Our worst-case constrained method matches or outperforms pure RL accuracy while retaining high fidelity.}
    \label{fig:pareto_front}
\end{figure}

\begin{table}[t]
\centering
\caption{Main distillation results across three settings. All metrics are reported in percentages (\%) and higher is better. Our method ($d=-\ln(1/4)$) achieves the highest Reasoning Success Rate (RSR) across all settings by breaking the trade-off between Final Answer Correctness (FAC) and Constraint Satisfaction (CS). Best results are in \textbf{bold}, second best are \underline{underlined}.}
\label{tab:main_results}
\resizebox{\textwidth}{!}{
\begin{tabular}{l ccc ccc ccc}
\toprule
& \multicolumn{3}{c}{\textbf{Qwen2.5 (MATH)}} & \multicolumn{3}{c}{\textbf{Llama-3.2 (MATH)}} & \multicolumn{3}{c}{\textbf{ Qwen2.5 (GSM-Sym.)}} \\
\cmidrule(lr){2-4} \cmidrule(lr){5-7} \cmidrule(lr){8-10}
\textbf{Method} & \textbf{RSR} & \textbf{FAC} & \textbf{CS} & \textbf{RSR} & \textbf{FAC} & \textbf{CS} & \textbf{RSR} & \textbf{FAC} & \textbf{CS} \\
\midrule
SFT on Teacher CoT & 24.66 & 27.02 & 78.54 & 13.64 & 17.48 & 57.72 & 55.78 & 65.9 & 83.46 \\
GKD & 24.00 & 25.84 & 85.18 & 9.34 & 15.24 & 65.54 & 57.58 & 66.06 & \underline{86.22} \\
MiniLLM & 24.22 & 26.28 & 85.68 & 11.08 & 15.20 & 61.88 & 55.28 & 63.96 & 83.88 \\
\midrule
GKD-GRPO $\lambda=10.0$ & 24.58 & 26.32 & 84.38 & 7.44 & 16.46 & \underline{66.88} & 56.56 & 65.70 & 83.70 \\
GKD-GRPO $\lambda=1.0$ & 24.32 & 26.40 & 85.14 & 9.60 & 17.16 & \textbf{67.76} & 57.18 & 65.78 & 83.98 \\
GKD-GRPO $\lambda=0.1$ & 26.56 & 28.90 & 84.66 & 11.98 & 18.56 & 66.82 & 58.66 & 66.68 & 84.32 \\
GKD-GRPO $\lambda=0.01$ & 29.42 & 32.86 & 78.46 & 12.48 & 17.92 & 59.58 & 61.04 & 70.68 & 83.80 \\
GKD-GRPO $\lambda=0.001$ & 29.48 & 34.62 & 61.48 & 11.44 & 18.08 & 34.14 & 64.02 & 74.08 & 0.00 \\
\midrule
GRPO & 27.54 & \underline{34.84} & 28.56 & 12.90 & \underline{18.78} & 14.28 & \underline{64.36} & \textbf{76.24} & 1.00 \\
\midrule
\textbf{Ours} $d=-\ln(1/4)$ & \textbf{30.00} & \textbf{34.86} & 81.30 & \textbf{14.12} & \textbf{20.54} & 56.94 & \textbf{65.82} & \underline{76.14} & 83.88 \\
\textbf{Ours} $d=-\ln(1/2)$ & \underline{29.76} & 33.82 & \textbf{86.58} & 13.34 & 18.08 & 62.58 & 63.06 & 72.96 & 60.44 \\
\textbf{Ours} $d=-\ln(3/4)$ & 26.36 & 28.80 & \underline{85.76} & \underline{13.92} & 16.9 & 64.96 & 58.96 & 67.40 & \textbf{86.30} \\
\midrule
\color{gray} Student & & 0.54 &  &  & 0.08 & &  & 0 & \\
\color{gray} Teacher & & 34.46 &  & & 32.08 & &  & 82.12 &  \\
\bottomrule
\end{tabular}
}
\end{table}

\subsection{RQ1: Bridging the Accuracy-Fidelity Trade-off}
A core premise of our work is that soft divergence constraints (like GKD-GRPO) invoke a fundamental compromise: to get high accuracy, one must abandon fidelity, leading to reward hacking. Figure~\ref{fig:pareto_front} and Table~\ref{tab:main_results} empirically support this.

In Table~\ref{tab:main_results}, looking at pure GRPO on Qwen+GSM-Symbolic, the model achieves a high FAC of 76.24\%, but suffers an almost zero Constraint Satisfaction rate (CS = 1\%). It has learned to game the final reward. The soft GKD-GRPO baseline attempts to fix this by interpolating with $\lambda$. However, reducing $\lambda$ to $0.001$ to recover accuracy (FAC 74.08\%) violently decreases CS to 0\%. Conversely, raising $\lambda$ to 0.1 fixes CS (84.32\%) but cripples task performance (FAC 66.68\%).

Our method ($d=-\ln(1/4)$) achieves a remarkable trade-off on the Pareto front on the 3 settings. By enforcing the constraint as a hard worst-case boundary during training, we guide the optimizer to valid regions even on the testing sets. As a result, our method matches the peak FAC of GRPO (76.14\% vs 76.24\%) while simultaneously holding CS high at 83.88\%. This remarkable dominance holds across settings: on Llama+MATH, we reach a peak FAC of 20.54\% (outperforming GRPO's 18.78\%) while quadrupling CS (56.94\% vs 14.28\%).

It is worth noting the sensitivity of the $d$ boundary. Table~\ref{tab:main_results} shows that an overly strict constraint ($d=-\ln(3/4)$) actively harms both FAC and RSR. Because the student has limited representational capacity compared to the teacher, forcing it to mimic the teacher perfectly prevents it from discovering simpler logic paths it can actually execute, establishing $d=-\ln(1/4)$ as the empirical sweet spot. Furthermore, we observe an expected train-test generalization gap: during training, our method satisfies the constraint almost strictly (CS $\approx$ 100\%), but domain shifts at test time reduce this to $\sim$80\%. Nonetheless, this inductive bias clearly prevents the catastrophic 0\% CS collapse seen in soft baselines optimized for accuracy.

\subsection{RQ2: Superior Logical Reasoning Quality}
\begin{wrapfigure}[16]{r}{0.4\textwidth}
    \includegraphics[width=1.0\linewidth]{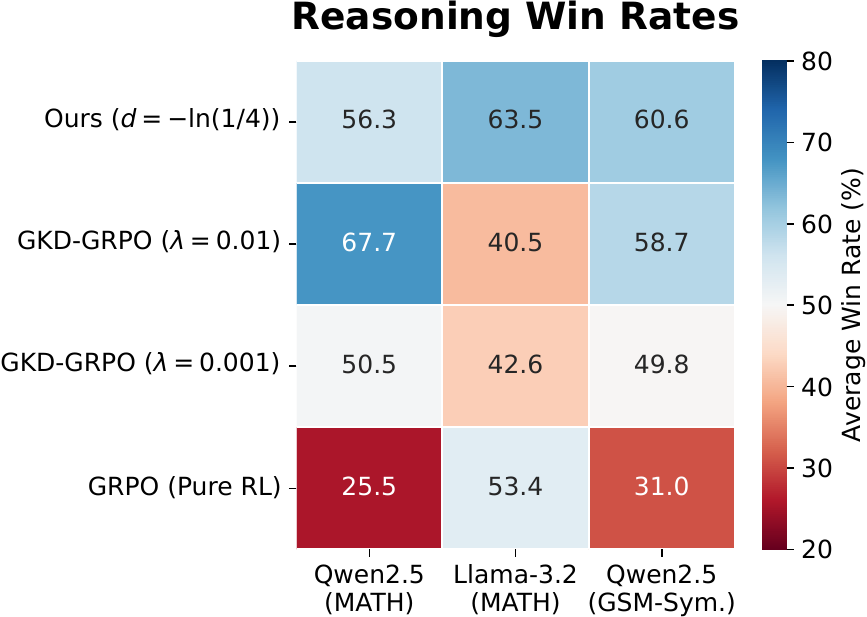}
    \caption{Average win rate (\%) evaluated by an LLM-as-a-Judge across the three distillation settings.}
    \label{fig:pairwise}
\end{wrapfigure}

Does suppressing constraint violations actually yield better reasoning?
To validate this rigorously, we conduct a strict Pairwise Judge evaluation. The judge prompt explicitly instructs the evaluating LLM to search for \textit{logical gaps and false leaps, even if the final answers are identical}. Figure~\ref{fig:pairwise} displays the overall win rate for each method across the datasets.

The results starkly confirm the "weakest link" compensation failure. Pure GRPO collapses under scrutiny: despite achieving high final answer correctness, its reasoning win rate plummets to 25.5\% on Qwen-MATH and 30.9\% on Qwen-GSM because the judge heavily penalizes its learned reward-hacks. Meanwhile, standard soft-regularization (GKD-GRPO $\lambda=0.01$) proves incredibly brittle: while it spikes to 67.7\% on Qwen-MATH, it completely breaks down on Llama-MATH (40.5\% win rate), showing that soft penalties are fundamentally sensitive to architecture and dataset dynamics.

In contrast, our worst-case constraint acts as a robust, invariant guardrail. Our method is the only approach to maintain a dominant, consistent win rate across all three discrete settings (56.2\%, 63.5\%, and 60.6\% respectively). The qualitative example in Figure~\ref{fig:qwen2_quali1_short} perfectly encapsulates this dynamic: GRPO learns an artificial rounding trick to secure the reward, which the judge flags, whereas our constraint forces the student to derive the mathematically sound, generalized path.

\subsection{RQ3: Empirical Validation of Anti-Compensation}

\begin{figure}[h]
    \centering
    \includegraphics[width=1.0\textwidth]{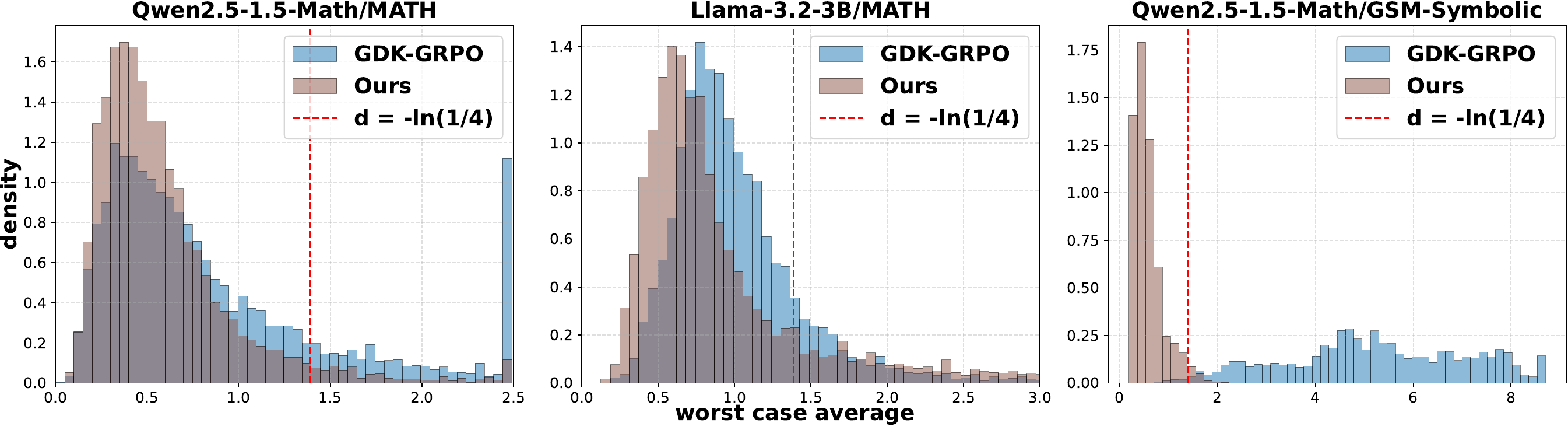}
    \caption{Distribution of the worst-case prefix average cost at test time. Soft baselines display extended "tails" of severe localized violations, empirically validating the compensation problem.}
    \label{fig:anti_compensation}
\end{figure}

We empirically validate this by analyzing the trained models at test time. For each generated test trajectory, we compute its specific worst-case prefix average cost $\max_{T} \frac{1}{T} \sum_{t=0}^{T-1} C_t$. 
Figure~\ref{fig:anti_compensation} plots the distribution of this metric (for legibility, we isolate our best reasoning variant $d=-\ln(1/4)$ against the highest-accuracy baseline GKD-GRPO $\lambda=0.001$). As predicted, the soft baseline exhibits a right-skewed "tail" (especially with the Qwen model). While their \textit{mean} sequence cost might be low, they can generate trajectories containing localized violations. In contrast, our method imposes a sharper cutoff at the boundary $d$, proving our method actively suppresses these compensations.

\section{Related Work}
\label{sec:related_work}

\paragraph{RL-Based Distillation and Reward Hacking.}
The traditional paradigm of LLM distillation aligns the student's output distribution with the teacher's via a divergence metric, typically reverse KL \citep{hinton2015distilling,sanh2020distilbertdistilledversionbert,minillm,zimmer2025mixture}. However, strictly mimicking the teacher ignores the downstream objective and can force the student into modeling complex distributions beyond its capacity \citep{zhang-etal-2025-towards-law}. Consequently, modern distillation increasingly leverages RL to incorporate verifiable task rewards (e.g., reaching the correct final answer) \citep{distillm,distillm2}.
While pure RL dramatically improves final task performance, it reliably induces \textit{reward hacking}: the student exploits logical shortcuts or hallucinatory leaps to reach the correct final answer without sound intermediate reasoning \citep{gudibande2024false}. To combat this, standard practices composite the RL reward with a soft KL regularizer against the teacher \citep{agarwal2024policy,rlhf,stiennon2022learningsummarizehumanfeedback,yang2024asymptoticslanguagemodelalignment}. %

\paragraph{Enhancing Reasoning Distillation.}
Because soft regularizers often fail to ensure rigorous reasoning, much of the recent literature relies on injecting dense, intermediate supervision. \textit{Process-aware} distillation supervises the student directly on the teacher's intermediate logical steps, transferring causal structure rather than just final outcomes \citep{distill-step-by-step, siked, chen2025step}. Other approaches include \textit{logit-aware} distillation, which utilizes attention or Bayesian heuristics to dynamically weight the distillation loss on "pivotal" tokens \citep{li2025bayeskd, li2024mixed, taskd-llm}, and \textit{knowledge-augmented} strategies that retrieve external information to patch knowledge gaps \citep{kang2023knowledge, 10.1145/3701551.3703577}. While effective, these methods introduce significant friction, requiring fine-grained step-level annotations, external process-reward models, or complex retrieval overhead. 

\paragraph{Constrained RL and State Augmentation.}
Preventing catastrophic deviations is the primary focus of Constrained RL \citep{10.5555/3305381.3305384, pmlr-v37-schulman15}. Framing LLM alignment as a strictly constrained problem offers strong theoretical guarantees but is computationally hostile. Standard continuous Constrained MDPs are solved via dual Lagrangian descents, which, at the scale of modern LLMs, require managing unstable dual variables across deep exploratory rollouts, incurring prohibitive compute, memory, and variance costs \citep{dasgupta-etal-2023-cost}.
Recent advances in safe RL avoid Lagrangian duality via \textit{state augmentation}, introducing frameworks like Saute \citep{sootla2022saute,sootla2022enhancing,10262328}, which append a remaining safety budget to the MDP state and enforce constraints via a reshaped hard-penalty reward.

\section{Conclusion}

In this work, we identified a fundamental flaw in RL-based LLM distillation: soft divergence penalties allow student models to mask severe logical errors through compensation. Recognizing that a reasoning chain is only as valid as its weakest link, we introduced a worst-case Constrained RL framework that enforces a strict boundary on the teacher log-likelihood along every prefix of the trajectory. To make this tractable at scale, we formulated an unaugmented MDP that strictly penalizes constraint violations without requiring test-time teacher access or unstable dual Lagrangian optimization. Extensive evaluations across multiple model families, coding and mathematical benchmarks demonstrate that our approach successfully breaks the accuracy-fidelity Pareto front. By strictly pruning invalid reasoning paths during exploration, our method matches the high final answer correctness of pure RL while drastically reducing constraint violations, ultimately yielding superior, trustworthy reasoning capabilities in compact models.

\bibliography{main}
\bibliographystyle{iclr2027_conference}

\clearpage
\appendix

\section{Derivations and Proofs}

\subsection{Proof of Proposition \ref{prop:anti_comp}}
\label{app:anti_comp}

\textbf{Part (a): Anti-Compensation} \\
Consider a generation trajectory $\tau = (\mathbf{s}_0, \mathbf{a}_0, \dots, \mathbf{s}_{T-1}, \mathbf{a}_{T-1})$ of length $T$, evaluated under a non-negative per-step cost $C_t \triangleq C(\mathbf{s}_t, \mathbf{a}_t) \geq 0$. Suppose the sequence satisfies a traditional average threshold (such as those induced by soft penalties evaluating the whole-sequence divergence), namely $\frac{1}{T} \sum_{t=0}^{T-1} C_t \leq d$.

To isolate the magnitude of a single-step hallucination, consider an adversarial scenario where the student perfectly mimics the teacher at all steps except for an arbitrary step $k$. Specifically, let $C_t = 0$ for all $t \neq k$.
To satisfy the soft average constraint, it is strictly required that:
\[
\frac{1}{T} \left( 0 + \dots + C_k + \dots + 0 \right) \leq d \implies C_k \leq T \cdot d.
\]
As $T$ grows (longer reasoning chains), the tolerated single-step deviation $C_k$ scales linearly with $T$. Thus, the maximum possible maximum single-step error $C_{\max} = \max_t C_t$ is bounded by $\mathcal{O}(T \cdot d)$, representing an unbounded localized hallucination completely masked by surrounding safe text.

Conversely, our worst-case average constraint requires that for \textit{all} prefix lengths $\tau \in \{1, \dots, T\}$, $\frac{1}{\tau} \sum_{t=0}^{\tau-1} C_t \leq d$.
At step $k$ (representing a prefix length of $k+1$ in 0-based indexing), evaluating the prefix sum up to $k$ yields:
\[
\frac{1}{k+1} \sum_{t=0}^{k} C_t \leq d.
\]
Since all $C_t \geq 0$, removing the terms prior to $k$ establishes a strict upper bound:
\[
\frac{1}{k+1} C_k \leq \frac{1}{k+1} \sum_{t=0}^{k} C_t \leq d \implies C_k \leq (k+1) \cdot d.
\]
Thus, the magnitude of a reasoning hallucination at step $k$ is tightly bounded by the prefix length $(k+1) \cdot d$, decisively preventing global averaging and the "hiding" of arbitrary magnitude logic breaks within long sequences.

\vspace{1em}
\textbf{Part (b): Non-Equivalence to Soft Lagrangian Regularization} \\
We are to prove there exists no constant scalar $\lambda \geq 0$ such that the unconstrained soft objective $J_{\text{soft}}(\pi) = \mathbb{E}_\pi [ \sum_{t} \gamma^t (R_t - \lambda C_t) ]$ shares an optimal policy with our constrained objective for all combinations of rewards and costs. We proceed by constructing a counterexample.

Consider a deterministic environment providing a final task reward at terminal step $T=2$. Let the budget $d=1$, with $\gamma=1$.
Let the optimization act over two possible sequences, $\tau_1$ and $\tau_2$.
\begin{itemize}
    \item \textbf{Sequence 1 (Faithful):} Task reward $R(\tau_1) = 1$. The costs are evenly distributed: $C_1 = 1, C_2 = 1$.
    \item \textbf{Sequence 2 (Reward Hack):} Task reward $R(\tau_2) = 1 + \epsilon$ (where $\epsilon > 0$). The costs are heavily skewed: $C_1 = 2, C_2 = 0$.
\end{itemize}

Evaluating under our worst-case constrained MDP ($n \to \infty$):
For $\tau_1$, prefixes average to $\frac{1}{1} \le 1$ and $\frac{1+1}{2} \le 1$. The trajectory is feasible, so its value is $V_{\text{con}}(\tau_1) = 1$.
For $\tau_2$, the first prefix averages to $\frac{3}{1} > 1$. The trajectory violates the constraint and earns the penalty, so $V_{\text{con}}(\tau_2) = -n < 0$.
The optimal constrained policy deterministically selects the faithful trajectory $\tau_1$ in our framework.

Now, evaluating under the generic soft Lagrangian objective:
The expected sum of costs for both trajectories is identical ($\sum C_t = 2$). Thus:
\begin{align*}
V_{\text{soft}}(\tau_1) &= 1 - \lambda (2) \\
V_{\text{soft}}(\tau_2) &= 1 + \epsilon - \lambda (2)
\end{align*}
For \textit{any} choice of $\lambda \ge 0$, we have $V_{\text{soft}}(\tau_2) > V_{\text{soft}}(\tau_1)$ by exactly $\epsilon$. The soft regularized objective will \textit{always} prefer the reward-hacked, locally hallucinatory trajectory $\tau_2$. Therefore, no scalar $\lambda$ can replicate the exact constraint boundary of our formulation, Q.E.D.

\subsection{Proof of Theorem \ref{thm:bellman}}
\label{app:bellman}

For this proof, we assume that the reward function is bounded, that we are in a discrete LLM distillation setting and the cost function is always positive. 
We first prove the Bellman equation holds for $\widehat{\mathcal{M}}_d^n$ by verifying conditions B1--B3 \citep{hernandez1992discrete}.
\begin{itemize}
    \item \textbf{B1 (Bounded and Measurable):} On feasible histories, $\hat{R}_n(\mathbf{s}, \mathbf{a}) = R(\mathbf{s}, \mathbf{a})$, which is uniformly bounded by assumption $R \in [0, R_{\max}]$. On infeasible histories (which by Lemma~\ref{lemma:absorbing} is an absorbing state), $\hat{R}_n(\mathbf{s}, \mathbf{a}) = -n$. The reward is piecewise bounded by $[-n, R_{\max}]$. In the discrete token mapping topology, all mappings are measurable.
    \item \textbf{B2 (Weak Continuity):} Since the joint space $\mathcal{S} \times \mathcal{A}$ of auto-regressive prefix sequences and discrete tokens is countable, any metric mapping naturally satisfies sequence convergence (all convergent sequences are eventually constant), thereby securing topological weak continuity.
    \item \textbf{B3 (Compactness):} The token vocabulary $\mathcal{A}$ is finite, meaning the action space is inherently bounded, closed, and thus compact.
\end{itemize}
Under these standard conditions and given $\gamma \in (0, 1]$, standard dynamic programming theory ensures the existence of a stationary Markov policy optimizing the Bellman equation for $\widehat{\mathcal{M}}_d^n$. The state $\mathbf{s}_t$ encodes the entire token history precisely; therefore, evaluating the feasibility bound $\max_{\tau \le T} \frac{1}{\tau} \sum C_t \le d$ preserves the Markov property.

\textbf{Monotone Value Convergence \& Almost Sure Safety:} \\
By construction in Eq.~\ref{eq:our_prob}, the penalty branch maps to $-n$. Thus, for any two penalty parameters $m > n$, we analytically hold $\hat{R}_m(\mathbf{s}, \mathbf{a}) \leq \hat{R}_n(\mathbf{s}, \mathbf{a})$ everywhere. Evaluating the discounted return over time, this injects a strict inequality $V_m^*(\mathbf{s}) \le V_n^*(\mathbf{s})$. As $\{\hat{V}_n^*(\mathbf{s})\}_{n\ge 1}$ is monotone non-increasing and trivially lower-bounded below on valid subsets, it must converge to a point-wise limit $\hat{V}_\infty^*$.

In the limiting target $\widehat{\mathcal{M}}_d^\infty$ (where $n \to \infty$), any trajectory passing through the absorbing infeasible boundary $\mathcal{S}_{\text{inf}}$ obtains a return of $-\infty$.
If it is given that the optimum policy $\pi^*$ for $\widehat{\mathcal{M}}_d^\infty$ achieves a strictly finite value, its probability mass entering the penalty region must be uniformly zero. Put simply:
\[
\mathbb{P}_{\pi^*}\!\Big(\exists\, T \ge 1 \text{ s.t. } \frac{1}{T}\sum_{t=0}^{T-1}C_t > d \Big) = 0.
\]
The complementary probability dictates that the sequence remains feasible:
\[
\mathbb{P}_{\pi^*}\!\Big(\max_{T\ge 1}\frac{1}{T}\sum_{t=0}^{T-1}C(\mathbf{s}_t, \mathbf{a}_t) \le d\Big) = 1,
\]
meaning it satisfies the constraint \textit{almost surely}.

\paragraph{Remark on the finite-value assumption.}
Theorem~\ref{thm:bellman} assumes the limiting MDP $\widehat{\mathcal{M}}_d^\infty$ admits an optimal policy with finite value; we argue this is a mild and verifiable condition rather than a technicality. 
A sufficient condition is the existence of \emph{any} feasible policy with finite expected return. 
Concretely, let $d$ be chosen at a high quantile of the worst-case prefix-average self-cost of correct-answer rollouts produced by an imitation (SFT) policy; by construction those rollouts are feasible, and since $R \in [0, R_{\max}]$ their value is finite, so the assumption holds. 
This yields a practical feasibility check: if no policy in the candidate pool can satisfy the constraint, the budget $d$ is set below the minimal achievable worst-case cost, \emph{every} policy incurs the $-\infty$ penalty in the limit, and the assumption fails by construction: the constraint itself is infeasible, and no method could satisfy it.
In other words, the assumption excludes only degenerate budgets.

\subsection{Policy Gradient Derivation}
\label{app:derive}

We derive the exact likelihood-ratio policy gradient given the discrete constraint MDP framework. Note that we treat the constrained token-level reward $\hat{R}_n(\mathbf{s}_k, \mathbf{a}_k)$ explicitly as a fixed scalar coefficient decoupled from $\theta$ (the student avoids differentiating through the frozen teacher mapping).

The objective is $J_n(\theta) = \mathbb{E}_{\pi_\theta} \left[ \sum_{t=0}^\infty \gamma^t \hat{R}_n(\mathbf{s}_t, \mathbf{a}_t) \right]$. Under standard REINFORCE \citep{sutton1999policy}:
\begin{align}
\label{eq:app_reinforce}
\nabla_\theta J_n(\theta)
&= \mathbb{E}_{\pi_\theta} \left[ \sum_{t=0}^{\infty} \gamma^t \, R_t^n \, \nabla_\theta \log \pi_\theta(\mathbf{a}_t \mid \mathbf{s}_t) \right], \\
\text{where} \quad R_t^n &\triangleq \sum_{k=t}^{\infty} \gamma^{k-t} \hat{R}_n(\mathbf{s}_k, \mathbf{a}_k) = \hat{R}_n(\mathbf{s}_t, \mathbf{a}_t) + \gamma R_{t+1}^n. \notag
\end{align}
We inject the recursive rollout function back into the gradient formulation to decompose it:
\begin{align}
\nabla_\theta J_n(\theta)
&= \mathbb{E}_{\pi_\theta} \left[ \sum_{t=0}^{\infty} \gamma^t \left( \hat{R}_n(\mathbf{s}_t, \mathbf{a}_t) + \gamma R_{t+1}^n \right) \nabla_\theta \log \pi_\theta(\mathbf{a}_t \mid \mathbf{s}_t) \right] \notag \\
&= \mathbb{E}_{\pi_\theta} \left[ \sum_{t=0}^{\infty} \gamma^t \hat{R}_n(\mathbf{s}_t, \mathbf{a}_t) \nabla_\theta \log \pi_\theta(\mathbf{a}_t \mid \mathbf{s}_t) \right] + \mathbb{E}_{\pi_\theta} \left[ \sum_{t=0}^{\infty} \gamma^{t+1} R_{t+1}^n \nabla_\theta \log \pi_\theta(\mathbf{a}_t \mid \mathbf{s}_t) \right]. \label{eq:decompose}
\end{align}

Extracting the first term (the \emph{single-step} component), we resolve the expectation explicitly over the token prediction step by conditioning on $\mathbf{s}_t$:
\begin{align}
\mathbb{E}_{\mathbf{a}_t \sim \pi_\theta(\cdot \mid \mathbf{s}_t)} \Big[ \hat{R}_n(\mathbf{s}_t, \mathbf{a}_t) \nabla_\theta \log \pi_\theta(\mathbf{a}_t \mid \mathbf{s}_t) \Big]
&= \sum_{\mathbf{a}} \pi_\theta(\mathbf{a} \mid \mathbf{s}_t) \hat{R}_n(\mathbf{s}_t, \mathbf{a}) \nabla_\theta \log \pi_\theta(\mathbf{a} \mid \mathbf{s}_t) \notag \\
&= \sum_{\mathbf{a}} \hat{R}_n(\mathbf{s}_t, \mathbf{a}) \nabla_\theta \pi_\theta(\mathbf{a} \mid \mathbf{s}_t) \notag \\
&= \nabla_\theta \sum_{\mathbf{a}} \pi_\theta(\mathbf{a} \mid \mathbf{s}_t) \hat{R}_n(\mathbf{s}_t, \mathbf{a}). \label{eq:exact_step}
\end{align}
Where we manipulated $\nabla_\theta \log \pi(x) = \frac{\nabla_\theta \pi(x)}{\pi(x)}$, and utilized the fact that, conditional on the prefix $\mathbf{s}_t$, $\hat{R}_n$ acts strictly as an external constant multiplier invariant to $\theta$. Substituting Eq.~\ref{eq:exact_step} back into Eq.~\ref{eq:decompose} produces the finalized, unbiased decomposed gradient exactly aligning with Section \ref{subsec:pg_opt}.

\subsection{Variance Reduction}
\label{app:variance}

\begin{theorem}[Per-Step Variance Reduction via Rao-Blackwellization]
Let $X_t = \hat{R}_n(\mathbf{s}_t, \mathbf{a}_t) \nabla_\theta \log \pi_\theta(\mathbf{a}_t \mid \mathbf{s}_t)$ be the standard REINFORCE gradient acting on the immediate reward at step $t$, assuming $\mathbb{E}[\|X_t\|^2] < \infty$, and let $Y_t = \sum_{\mathbf{a}} \pi_\theta(\mathbf{a}\mid \mathbf{s}_t) \hat{R}_{n}(\mathbf{s}_t,\mathbf{a}) \nabla_\theta \log \pi_\theta(\mathbf{a} \mid \mathbf{s}_t)$ be its exact-expectation counterpart within $(\nabla_\theta J_n)_{\textsc{single}}$. By the law of total covariance, conditioning on the prefix $\mathbf{s}_t$ ensures that the trace of the covariance matrices satisfies:
\begin{equation}
\mathrm{Tr}(\mathrm{Cov}(Y_t)) \leq \mathrm{Tr}(\mathrm{Cov}(X_t)) \nonumber
\end{equation}
with strict inequality whenever $\mathbb{P}(\mathrm{Cov}(X_t \mid \mathbf{s}_t) \neq \mathbf{0}) > 0$.
\label{thm:variance}
\end{theorem}

We wish to prove that analytically computing the exact expectation of the immediate reward step strictly reduces the total variance of the gradient estimator via Rao-Blackwellization \citep{JMLR:v21:19-346}.

Let the standard REINFORCE gradient component for the immediate reward at step $t$ be defined as the stochastic random vector:
$$ X_t = \hat{R}_n(\mathbf{s}_t, \mathbf{a}_t) \nabla_\theta \log \pi_\theta(\mathbf{a}_t \mid \mathbf{s}_t) $$
where the action $\mathbf{a}_t$ is sampled from the student policy $\pi_\theta(\cdot \mid \mathbf{s}_t)$. 

Instead of relying on a single Monte Carlo sample, our decomposed estimator evaluates the conditional expectation of $X_t$ over the entire vocabulary, given the current prefix $\mathbf{s}_t$:
$$ Y_t = \mathbb{E}_{\mathbf{a}_t \sim \pi_\theta(\cdot \mid \mathbf{s}_t)} [X_t \mid \mathbf{s}_t] = \sum_{\mathbf{a}} \pi_\theta(\mathbf{a} \mid \mathbf{s}_t) \hat{R}_n(\mathbf{s}_t, \mathbf{a}) \nabla_\theta \log \pi_\theta(\mathbf{a} \mid \mathbf{s}_t) $$
Note that $Y_t$ exactly matches the $t$-th term of our decomposed $(\nabla_\theta J_n)_{\textsc{single}}$ construction, as derived in Appendix \ref{app:derive}.

To compare the variance of these two estimators, we apply the Law of Total Covariance to the vector $X_t$ conditioned on the state $\mathbf{s}_t$:
\begin{align}
\mathrm{Cov}(X_t) &= \mathrm{Cov}(\mathbb{E}[X_t \mid \mathbf{s}_t]) + \mathbb{E}[\mathrm{Cov}(X_t \mid \mathbf{s}_t)] \notag \\
\mathrm{Cov}(X_t) &= \mathrm{Cov}(Y_t) + \mathbb{E}[\mathrm{Cov}(X_t \mid \mathbf{s}_t)]. \label{eq:tot_var}
\end{align}

Taking the trace of both sides to measure total variance, and noting that the trace of a positive semi-definite covariance matrix is non-negative, by linearity of expectation we have:
$$ \mathrm{Tr}(\mathbb{E}[\mathrm{Cov}(X_t \mid \mathbf{s}_t)]) = \mathbb{E}[\mathrm{Tr}(\mathrm{Cov}(X_t \mid \mathbf{s}_t))] \geq 0. $$
Subtracting this yields the fundamental inequality:
$$
\mathrm{Tr}(\mathrm{Cov}(Y_t)) \leq \mathrm{Tr}(\mathrm{Cov}(X_t)).
$$

Strict inequality holds if and only if $\mathbb{P}(\mathrm{Cov}(X_t \mid \mathbf{s}_t) \neq \mathbf{0}) > 0$. This condition is satisfied as long as the random vector $X_t \mid \mathbf{s}_t$ is not almost surely constant. In the context of our MDP, this requires the student policy $\pi_\theta(\cdot \mid \mathbf{s}_t)$ to be non-degenerate (i.e., it assigns non-zero probability to at least two actions) and for those actions to yield divergent values for the gradient vector $\hat{R}_n(\mathbf{s}_t, \mathbf{a}) \nabla_\theta \log \pi_\theta(\mathbf{a} \mid \mathbf{s}_t)$. 

\textbf{Practical Significance in our MDP:} While Rao-Blackwellization is a generic statistical tool \citep{JMLR:v21:19-346}, computing the exact expectation over an LLM vocabulary (often 50K-150K tokens) is typically intractable. It is uniquely unlocked here because evaluating $\hat{R}_n(\mathbf{s}_t, \mathbf{a})$ for all possible next tokens requires only a single forward pass of the teacher to obtain the log-probabilities, followed by a deterministic prefix-feasibility check. Furthermore, in standard unconstrained RL, the immediate reward $R(\mathbf{s}_t, \mathbf{a}_t)$ is often zero during intermediate reasoning steps, making $X_t = \mathbf{0}$ and rendering this decomposition trivial. However, in our worst-case constrained formulation, Lemma \ref{lemma:absorbing} dictates that a constraint violation immediately triggers a massive penalty $\hat{R}_n = -n$. This creates an extreme "cliff" in the reward landscape at step $t$, injecting large variance into the standard sampled estimator $X_t$. By analytically integrating over the vocabulary to compute $Y_t$, we perfectly smooth over this cliff, yielding a highly stable, low-variance gradient signal that safely steers the policy.

\section{Algorithm and Implementation Details}
\label{sec:app_algo}

We will open-source our code upon acceptance.

\subsection{Reward Function Design}
For mathematical reasoning tasks, we rely on rule-based, deterministic evaluators (e.g., symbolic math matching) to assign binary rewards based on final answer correctness:
\begin{equation}
R(\mathbf{s}_T, \mathbf{a}_T) = \begin{cases}
1.0 & \text{if the extracted final answer is correct} \\
0.0 & \text{if the extracted final answer is incorrect}
\end{cases}
\end{equation}
The reward is entirely sparse and only assigned at the terminal step of each trajectory when the complete solution inside the \texttt{\textbackslash boxed\{\}} tag is generated.

\subsection{Baselines Description}
To ensure fair and rigorous comparisons, all reinforcement learning baselines, including our own, are implemented on top of the \textbf{GRPO} \citep{grpo} framework, sharing the same group-relative advantage estimation and PPO-style clipping mechanics.
\begin{itemize}[leftmargin=1.5em]
    \item \textbf{GRPO:} Optimizes purely for the binary task reward $R$ without any teacher supervision. It tends to reward-hack by discovering flawed reasoning chains that accidentally yield the correct final token.
    \item \textbf{GKD \citep{agarwal2024policy}:} A pure distillation baseline. It optimizes entirely to minimize the divergence (maximize teacher likelihood) without observing the task reward $R$, resulting in high constraint satisfaction but low final accuracy.
    \item \textbf{Mini-LLM \citep{minillm}:} A sequence-level distillation method that separates single-step and long-term divergence gradients \citep{tang2025few}. We adapt it over GRPO for consistency. We also sample trajectories exclusively with the student policy for fair comparison.
    \item \textbf{GKD-GRPO \citep{agarwal2024policy}:} The standard soft Lagrangian relaxation, optimizing $R - \lambda \cdot \text{KL}(\pi||\mu)$. We conduct a thorough grid search over $\lambda \in \{10, 1.0, 0.1, 0.01, 0.001\}$ to map out its entire Pareto frontier.
    \item \textbf{SFT on Teacher CoT:} Off-policy sequence-level distillation. We sample teacher chain-of-thought trajectories on the training set, keep only those with correct final answers, and fine-tune the student on them with the standard next-token loss for 4 epochs. The learning rate is the same as other baselines $1 \times 10^{-5}$.
\end{itemize}

\subsection{Hyperparameter Settings}
We maintained consistent hyperparameter settings across all methods to isolate the effect of the optimization objectives:
\begin{itemize}[leftmargin=1.5em]
    \item \textbf{Batch Size:} 64 responses per update (8 distinct questions $\times$ 8 sampled responses per question).
    \item \textbf{Learning Rate:} $1\times 10^{-5}$ with a constant schedule.
    \item \textbf{Optimizer:} AdamW.
    \item \textbf{Discount Factor:} $\gamma = 1.0$ (standard for finite-horizon LLM generation).
    \item \textbf{Training precision:} bfloat16.
    \item \textbf{Training Epochs:} 20 epochs across all models.
    \item \textbf{Penalty ($n$):} For our method, the terminal violation penalty was set to $n = 20$, providing a harsh boundary signal well above the maximum positive task reward of $1.0$.
    \item \textbf{Constraint Threshold ($d$):} We swept our worst-case average bound across $d \in \{-\ln(3/4), -\ln(1/2), -\ln(1/4)\}$. Since $d$ limits the average negative log-likelihood, $-\ln(1/4)$ roughly translates to demanding the teacher assigns at least a 25\% worst average fractional probability to the student's reasoning sequence thus far.
\end{itemize}

\textbf{Llama-3.2 Bootstrap:} Llama-3.2-3B exhibited severe training instability during pure GRPO training due to an extremely poor initial zero-shot performance on MATH (preventing any positive reward signals from being discovered). As a standard remedy, to bootstrap all evaluated methods equally, we applied a brief warm-up phase of pure distillation (using GKD) for the first 3 epochs before engaging the respective RL/Constrained objectives. No such bootstrap was required for Qwen models.

\subsection{Computational Resources and Training Time}
Training takes fewer than 23 hours on a single accelerator per method. Because the forward generation and backward gradient passes dominate the wall-clock time, the overhead of evaluating the teacher model's likelihood (required for GKD, Mini-LLM, GKD-GRPO, and Our method) during RL rollouts adds a small overhead to overall training time. Consequently, all methods are directly comparable in terms of computational training cost, except the one without a teacher (GRPO) that runs in around 16 hours.

\subsection{Evaluating Reasoning Quality (LLM-as-a-Judge)}
To rigorously assess the logical validity of generated solutions beyond verifiable final answers, we employ an LLM-as-a-Judge protocol \citep{zheng2023judging}. Following recent literature evaluating mathematical reasoning, we utilize \textit{DeepSeek-R1-Distill-Qwen-32B} as our evaluator. We run two distinct types of evaluations: a single-answer audit to compute the Reasoning Success Rate (RSR), and a head-to-head Pairwise Comparison to validate relative superiority.
Note that for the pairwise comparison, we use both orders of the answer. To conclude if one is winning, it has to win in both positions.

\subsubsection{Reasoning Success Rate (RSR) Prompt}
The RSR metric aims to identify any hallucination, invalid inference, or broken logical chain in a single trajectory. A generation is marked as a "Success" if achieving the correct final answer \textit{and} the judge returns a \texttt{No Error} verdict.

\begin{quote}
\small
\noindent\texttt{You are an expert auditor evaluating the reasoning correctness of an AI model's solution to a mathematical problem.}\\

\noindent\texttt{Your task:}\\
\texttt{- Determine whether the solution's reasoning contains ANY error.}\\
\texttt{- This is independent of whether the final answer is correct or not.}\\
\texttt{- An "error" includes: incorrect algebra/arithmetic, invalid inference, misuse of a theorem/definition, contradiction, or an unjustified assumption that changes the problem.}\\
\texttt{- Do NOT count: harmless typos, purely stylistic issues, or alternative valid approaches.}\\

\noindent\texttt{Instructions:}\\
\texttt{1) Read the problem and the solution.}\\
\texttt{2) Check the reasoning step-by-step for validity.}\\
\texttt{3) If you find an error, identify the FIRST point where the reasoning becomes invalid.}\\
\texttt{4) If the reasoning is too incomplete or ambiguous to evaluate, use "Unclear".}\\

\noindent\texttt{Problem:}\\
\texttt{```problem}\\
\texttt{\{prompt\}}\\
\texttt{```}\\
\texttt{Solution:}\\
\texttt{```solution}\\
\texttt{\{answer\}}\\
\texttt{```}\\

\noindent\texttt{Respond in the following exact format (one item per line):}\\
\texttt{Verdict: [No Error / Has Error / Unclear]}\\
\texttt{Severity: [Critical / Non-critical / N/A]}\\
\texttt{First\_Error\_Location: [Brief reference to the earliest wrong step, or N/A]}\\
\texttt{Explanation: [2-4 sentences explaining why you chose the verdict, focusing on the key reasoning issue(s)]}
\end{quote}

\subsubsection{Pairwise Ranker Prompt}
For the Pairwise evaluation, the judge is presented with two solutions from different models that both achieved the correct final answer. It is forced to choose which model provided structurally and logically superior reasoning.
We validate the judge against human annotation in Appendix~\ref{app:judge_reliability}.

\begin{quote}
\small
\noindent\texttt{You are an expert judge evaluating the reasoning quality of two AI models' responses to a mathematical problem.}\\

\noindent\texttt{Please compare the reasoning steps, clarity, and mathematical rigor of both responses.
You do not need to provide the answer to the problem itself as both responses arrive at the correct answer, so focus on:}\\
\texttt{1. Clarity and structure of reasoning steps}\\
\texttt{2. Mathematical rigor and correctness of intermediate steps}\\
\texttt{3. Explanation quality and logical flow}\\
\texttt{4. Completeness of the solution process}\\

\noindent\texttt{Problem:}\\
\texttt{```problem}\\
\texttt{\{prompt\}}\\
\texttt{```}\\
\texttt{Response A:}\\
\texttt{```response}\\
\texttt{\{answer\_a\}}\\
\texttt{```}\\
\texttt{Response B:}\\
\texttt{```response}\\
\texttt{\{answer\_b\}}\\
\texttt{```}\\

\noindent\texttt{Please provide an explanation of your reasoning in 2-3 sentences focusing on the key differences in reasoning quality.}\\

\noindent\texttt{Then provide your judgment as one of: "A wins", "B wins", or "Tie".}\\

\noindent\texttt{Format your response as:}\\
\texttt{Explanation: [Your explanation here]}\\
\texttt{Verdict: [A wins/B wins/Tie]}
\end{quote}

\section{Code Generation Experiment}
\label{app:code_gen}
To test generality beyond mathematics, we distill \textit{Qwen2.5-Coder-0.5B} (base) from \textit{Qwen2.5-Coder-7B-Instruct}, a third model pair, on problems from MBPP (train+validation) and APPS (introductory and interview levels) with a binary all-tests-pass reward, and evaluate pass@1 on the MBPP test set. 

\paragraph{Models.} Teacher: \textit{Qwen2.5-Coder-7B-Instruct}; student: \textit{Qwen2.5-Coder-0.5B} (base, non-instruct). This is a third model pair, distinct from both math settings, with a $14\times$ parameter ratio.

\paragraph{Data.} Training uses ${\sim}4$K problems: MBPP train+validation (464 problems) plus APPS at the introductory and interview difficulty levels, filtered for problem statements under 3{,}000 characters. Evaluation is pass@1 on the MBPP test set (500 problems). Prompts contain the problem statement and, following the MBPP convention, the unit-test assertions; for APPS, starter code when present.

\paragraph{Reward.} Binary: 1.0 if all unit tests pass, 0.0 otherwise. MBPP assertions are appended to the generated code and executed as a script; APPS problems are evaluated either by calling the target function on the provided inputs (when a function name is specified) or by piping inputs through stdin and comparing stdout. Each candidate runs in an isolated \texttt{python3} subprocess with a 5-second timeout.

\paragraph{Training.} Identical GRPO-based configuration to the math settings (batch of 8 prompts $\times$ 8 generations, 512-token completions, temperature 1.0), with 15 epochs over the training set. The $\lambda$ grid for GKD-GRPO and the $d$ grid for ours follow the math settings; we report the strongest configuration per method alongside pure GRPO.

\begin{table}[h]
\centering
\caption{Code generation on the MBPP test set (500 problems, \%). The prompt contains the unit tests, making pass@1 easy to game: GRPO inflates pass@1 to 92.0\% by copying them into the generated code (RSR 1.8\%). Ours achieves the highest RSR and smallest FAC$-$RSR gap.}
\label{tab:code_gen}
\begin{tabular}{l ccc}
\toprule
\textbf{Method} & \textbf{RSR} & \textbf{FAC (Pass@1)} & \textbf{FAC$-$RSR} \\
\midrule
GRPO & 1.8 & \textbf{92.0} & 90.2 \\
GKD-GRPO $\lambda=0.001$ & 14.4 & 74.4 & 60.0 \\
GKD-GRPO $\lambda=0.01$ & 34.0 & 50.4 & 16.4 \\
\textbf{Ours} $d=-\ln(1/4)$ & \textbf{34.2} & 45.6 & \textbf{11.4} \\
\bottomrule
\end{tabular}
\end{table}

Table~\ref{tab:code_gen} shows GRPO exploits the reward brutally: it simply copies the unit tests visible in the prompt into the generated code (pass@1 92.0\%, RSR 1.8\%). Soft regularization trades this off along the same Pareto axis as in math, while ours obtains the highest RSR (34.2\%) with the smallest FAC$-$RSR gap (11.4 points), confirming that the worst-case constraint prevents reward hacking in a new domain, model pair, and reward type.
This is a clear example showing that trusting FAC alone is not a good idea as it can produce worst models.

\section{Details on Constraint Type Analysis (Figure~\ref{fig:constraint_type})}
\label{app:constraint_analysis}
To empirically validate the structural advantages of our worst-case average constraint over the traditional Saute cumulative-sum constraint (discussed in Section~\ref{sec:worst_case_rl}), we conducted a retrospective test-time analysis using generated trajectories.

Using the Qwen2.5-1.5B student model, we oversample 256 trajectories per question. Because the sampler explores without constraints, this set features a wide variety of both logically sound solutions and reward-hacked solutions, providing a diverse distribution of teacher costs $C(\mathbf{s}_t, \mathbf{a}_t) = -\log \mu(\mathbf{a}_t | \mathbf{s}_t)$.

For each trajectory, we computed two properties:
\begin{enumerate}[leftmargin=1.5em]
    \item \textbf{Cumulative Budget:} $S_{\text{sum}} = \sum_{t=0}^{T-1} C_t$
    \item \textbf{Worst-Case Average Budget:} $S_{\text{wc}} = \max_{\tau \le T} \frac{1}{\tau} \sum_{t=0}^{\tau-1} C_t$
\end{enumerate}

We then simulated the effect of the two constraint mechanisms by treating them as test-time filters. By sweeping the budget parameters $d_{\text{sum}}$ and $d_{\text{wc}}$ across a fine-grained grid (the X-axis of Figure~\ref{fig:constraint_type}): from $0$ to $-3$ for the worst-case average and from $0$ to $-200$ for the cumulative sum (the scales differ because the cumulative cost grows with sequence length while the average is length-normalized), we "accepted" only the trajectories that satisfied the respective constraint bounds ($S \le d$).

For each subset of accepted trajectories, we calculated the overall Reasoning Success Rate (RSR). The resulting plot demonstrates that at its optimal budget threshold, the worst-case average filter achieves a strictly higher RSR than the optimal cumulative sum filter. This occurs because the cumulative sum indiscriminately rejects long, detailed, valid reasoning chains (which accumulate high total costs simply due to sequence length), whereas the worst-case average constraint specifically targets and isolates trajectories containing concentrated logical hallucinations, successfully filtering out reward-hacked responses regardless of their token length.

\paragraph{Alternative constraint parameterizations.} Beyond the cumulative sum, one might consider (i) a {per-step} constraint $C_t \le d\,\forall t$, or (ii) a \emph{sliding-window} average over the last $k$ steps. 

The per-step form is strictly harder to satisfy than the prefix average and degenerates for teachers with high entropy on formatting tokens (e.g., whitespace or LaTeX delimiters), whose individual costs routinely exceed any useful threshold $d$; it therefore conflates stylistic uncertainty with logical hallucination. 

The sliding-window average mitigates this but loses the absorbing (prefix-monotone) structure that makes the worst-case prefix average a valid safety budget in the Saute sense: once a prefix average is violated, no future tokens can repair it, which is precisely what enables our state-augmentation-free terminal penalty (Section~\ref{sec:worst_case_rl}). A windowed statistic can be repaired by subsequent low-cost tokens, reintroducing exactly the compensation failure the constraint is designed to prevent. 

We therefore retain the prefix average, which interpolates between the two extremes: length-invariant like the windowed form, absorbing like the per-step form.

\section{Trajectory-Level Validation of the Likelihood Proxy}
\label{app:proxy_validation}

Our constraint assumes that worst-case teacher likelihood tracks reasoning validity.
We test this directly on all test trajectories of 12 training runs spanning GRPO, GKD-GRPO across the full $\lambda$ grid, and ours across the $d$ grid (Qwen-MATH setting; 5{,}000 generations per run, 60{,}000 trajectories in total).
For each trajectory we record its worst-case prefix-average teacher log-likelihood and its RSR label (correct final answer and judge verdict ``No Error'').
Additionally, instead of considering only the teacher likelihood for the definition of the cost $C_t$, we also tried to replace it with per-token reverse and forward KL divergences.

\begin{table}[h]
\centering                                                                                                           
\caption{Pooled Spearman rank correlation between per-trajectory fidelity statistics and RSR (divergence costs are negated so that higher always means higher teacher fidelity).}
\label{tab:proxy_corr}
\begin{tabular}{l c}                                                                                                 
\toprule
\textbf{Fidelity statistic (per trajectory)} & \textbf{Spearman $\rho$ with RSR} \\
\midrule
Worst-case prefix avg.\ teacher logp (our constraint) & $+0.231$ \\
Worst-case prefix avg.\ reverse KL & $+0.224$ \\
Worst-case prefix avg.\ forward KL & $+0.186$ \\
\bottomrule
\end{tabular}
\end{table}

Table~\ref{tab:proxy_corr} shows the correlation is positive and significant, and it is positive in each of the 12 runs individually (per-run $\rho \in [0.13, 0.30]$).
Concretely, pooling trajectories into quartiles of worst-case log-likelihood, RSR rises monotonically from 30.1\% (lowest quartile) to 59.6\% (highest), and FAC from 35.9\% to 63.0\%. 
Two conclusions follow. 
First, the proxy assumption holds: trajectories that deviate sharply from the teacher at some prefix are substantially less likely to contain valid reasoning.
Second, replacing the teacher NLL with a forward KL or reverse KL per-token divergence would not improve the proxy.

Among answer-correct trajectories, judge-flagged (reward-hacked) ones have significantly lower worst-case log-likelihood than genuinely correct ones (mean $-1.09$ vs.\ $-0.92$ for GRPO, Mann--Whitney $p<10^{-6}$; pooled AUC $0.63$), confirming the constraint quantity carries signal specifically about reasoning failures rather than stylistic deviation.

\paragraph{Post-hoc threshold selection.} 
Because the constraint is a threshold on a stored per-trajectory quantity, its effect can be simulated post hoc on validation rollouts without retraining: treating the worst-case log-likelihood as a filter, tighter thresholds trade coverage for reasoning quality monotonically. This provides an explicit coverage--quality dial that $\lambda$, which changes the training objective itself, does not.

\section{Judge Reliability: Human Evaluation}
\label{app:judge_reliability}
To calibrate the LLM-as-a-Judge protocol used for RSR, we manually audited 60 judge verdicts, stratified across methods (GRPO, GKD-GRPO, ours), datasets (MATH, GSM-Symbolic), and verdict types (flagged and unflagged), with the annotator blinded to the method. The human verdicts agree with the judge on all 60 cases (100\% agreement). We note two systematic judge tendencies observed during piloting (and mitigated in the final prompt): a bias toward flagging solutions that reach the correct answer through unusual-but-valid shortcuts, and occasional leniency toward compact arithmetic slips; both were rare in the audited sample. The pairwise protocol additionally requires a method to win in both presentation orders, which filters position bias.

\section{Limitations and Future Work}
\label{app:limitations}
We acknowledge several limitations of our approach.

\textbf{Teacher likelihood as a cost proxy.} Our worst-case constraint relies on teacher log-likelihood, which correlates strongly with, but does not perfectly capture, reasoning validity. A step that is logically valid but stylistically unusual may incur a high cost. Exploring richer cost functions (e.g., learned process reward models) is a promising direction.

\textbf{Tokenizer alignment requirement.} Our method requires that the student and teacher models share a compatible tokenizer, as the per-step cost $C(\mathbf{s}_t, \mathbf{a}_t) = -\log \mu(\mathbf{a}_t | \mathbf{s}_t)$ must be evaluated token-wise at positions consistent between both models. This constraint is naturally satisfied when distilling within the same model family (e.g., Qwen2.5-7B $\to$ Qwen2.5-1.5B, Llama-3.2-11B $\to$ Llama-3.2-3B), as used in our experiments. Distilling across families with incompatible vocabularies would require either re-tokenization strategies, cross-tokenizer alignment techniques, or alternative cost formulations based on sequence-level or semantic similarity, which we leave to future work.

\textbf{Distribution shift at test-time.} Our theoretical guarantees hold under the training distribution. Extending formal feasibility guarantees to OOD generalization remains an open problem.

\textbf{Budget selection.} While $d$ is more interpretable than a Lagrangian multiplier $\lambda$ (it relates directly to teacher probability), optimal budget selection still requires some empirical tuning. 
However, unlike $\lambda$, the threshold acts on a stored per-trajectory quantity and can therefore be swept post hoc on validation rollouts without retraining, trading accepted-trajectory coverage against RSR (Section~\ref{app:proxy_validation}).
Adaptive budget schedules could further improve this.

This tuning requirement is not unique to our method: soft baselines must equally tune their Lagrangian multiplier $\lambda$, which typically requires sweeping several orders of magnitude and full retraining for each value.
It is inherent to fidelity--accuracy trade-off methods.

\textbf{Domain scope.} Our experiments focus on mathematical reasoning and code generation. Extending the worst-case constraint framework to broader reasoning tasks (theorem proving, multi-hop QA, agentic framework) is an important direction for future work.

\section{Generated Answers}
\label{sec:app_prompts}

We report more examples with reward hacking below:

\begin{figure}[h]
    \centering 
    \begin{tikzpicture}[
        boxa/.style={
            draw,
            rectangle,
            rounded corners,
            fill=green!5,
            text width=0.48\linewidth,
            minimum height=10.6cm,
            align=left,
            font=\scriptsize,
            text height=8.4cm,
            execute at begin node={\vbox\bgroup\raggedright},
            execute at end node={\vfil\egroup},
        },
        boxb/.style={
            draw,
            rectangle,
            rounded corners,
            fill=red!5,
            text width=0.48\linewidth,
            align=left,
            font=\scriptsize,
        },
        box_hoz/.style={
            draw,
            rectangle,
            rounded corners,
            fill=blue!5,
            text width=0.98\linewidth,
            minimum height=1cm,
            align=left,
            font=\scriptsize,
            inner sep=4pt
        },
        node distance=0.1cm
    ]
        
    \node (box2) [boxa] {
        \textbf{Ours}:\\[0.2cm]
        To determine what percentage of the whale's body length the combined length of the remoras is, we need to follow these steps:\par
        \textbf{1. Convert the length of the remoras from inches to feet:}
        \[ 45 \text{ inches} \times \frac{1 \text{ foot}}{12 \text{ inches}} = 3.75 \text{ feet} \]
        So, each remora is 3.75 feet long.\par
        \textbf{2. Calculate the total length of the remoras:}
        \[
           4 \text{ remoras} \times 3.75 \text{ feet per remora} = 15 \text{ feet}
        \]
        \textbf{3. Determine the total length of the whale:}
        \[
           300 \text{ feet}
        \]\par
        \textbf{4. Calculate the percentage of the whale's body length that the combined length of the remoras represents:}
        \[
           \left( \frac{15 \text{ feet}}{300 \text{ feet}} \right) \times 100\%
        \]
        \textbf{5. Simplify the fraction:}
        \[
           \frac{15}{300} = 0.05
        \]\par
        \textbf{6. Convert the decimal to a percentage:}
        \[
           0.05 \times 100\% = 5\%
        \]
        Therefore, the combined length of the remoras is \(\boxed{5}\) percent of the whale's body length.
    };
    
    \node (box3) [boxb, right=of box2] {
        \textbf{GRPO}:\\[0.2cm]
Let's break down the problem step by step to find the percentage of the whale's body length that the combined length of the remoras represents.\par
First, we need to convert all units to the same unit. Let's use feet for this calculation.\par
The length of each remora is given as 45 inches. Since there are 12 inches in a foot, the length of each remora in feet is:\par
\[ \frac{45}{12} = 3.75 \text{ feet} \]\par
Since there are 4 remoras, the total length of the remoras is:\par
\[ 4 \times 3.75 = 15 \text{ feet} \]\par
Next, we add the length of the whale to the combined length of the remoras to get the total length of the whale:\par
\[ 300 + 15 = 315 \text{ feet} \]\par
Now, we need to find what percentage the combined length of the remoras is of the whale's body length. This can be calculated using the formula for percentage:\par
\[ \left( \frac{15}{315} \right) \times 100 \]\par
Simplifying the fraction inside the percentage formula:\par
\[ \frac{15}{315} = \frac{1}{21} \]\par
\[ \left( \frac{1}{21} \right) \times 100 = \frac{100}{21} \approx 4.7619 \]\par
Rounding to the nearest whole number, we get: \[ \boxed{5} \]
    };
    \coordinate (midpoint_2_3) at ($(box2.center)!0.5!(box3.center)$);

    \node (box1) [box_hoz, above=of midpoint_2_3,yshift=5.3cm] {
        \textbf{Question}:\\
        Leilani saw a 300-foot whale with 4 45-inch remoras attached to it. What percentage of the whale's body length is the combined length of the remoras?
    };
    
    \node (box4) [box_hoz, below=of $(box2.center)!0.5!(box3.center)$,yshift=-5.3cm] {
        \textbf{Judge}:\\
        Response A correctly calculates the percentage by comparing the remoras' total length to the whale's length without adding them, ensuring accuracy. Response B incorrectly adds the remoras' length to the whale's, leading to a flawed calculation.\par
        Verdict: A wins
    };
    
    \end{tikzpicture}
    \caption{Example of generated answer with Qwen2.5-1.5B after distillation.}
    \label{fig:qwen2_quali1}
\end{figure}

\begin{figure}[h]
    \centering 
    \begin{tikzpicture}[
        boxa/.style={
            draw,
            rectangle,
            rounded corners,
            fill=green!5,
            text width=0.48\linewidth,
            align=left,
            font=\scriptsize,
        },
        boxb/.style={
            draw,
            rectangle,
            rounded corners,
            fill=red!5,
            minimum height=10.8cm,
            text width=0.48\linewidth,
            align=left,
            font=\scriptsize,
            execute at begin node={\vbox\bgroup\raggedright},
            execute at end node={\vfil\egroup},
            text height=6.6cm,
        },
        box_hoz/.style={
            draw,
            rectangle,
            rounded corners,
            fill=blue!5,
            text width=0.98\linewidth,
            minimum height=1cm,
            align=left,
            font=\scriptsize,
            inner sep=4pt
        },
        node distance=0.1cm
    ]
        
    \node (box2) [boxa] {
        \textbf{Ours}:\\[0.2cm]
        To determine what percentage of the whale's body length the combined length of the remoras is, we need to follow these steps:\par
        1. \textbf{Convert the length of the remoras from inches to feet:}\par   
        - Each remora is 12 inches long.\par
        - There are 12 inches in a foot.\par
        - Therefore, the length of each remora in feet is:\par     
        \[  \frac{12 \text{ inches}}{12 \text{ inches per foot}} = 1 \text{ foot}\]  
        - Since there are 2 remoras, the total length of the remoras in feet is:   
        \[2 \text{ remoras} \times 1 \text{ foot per remora} = 2 \text{ feet}  \]
        2. \textbf{Calculate the percentage of the whale's body length:}\par   
        - The whale's body length is 200 feet.\par   
        - The combined length of the remoras is 2 feet.\par   
        - To find the percentage, we use the formula:\par
        \[ \text{Percentage} = \left( \frac{\text{Length of remoras}}{\text{Whale's body length}} \right) \times 100 \]
        - Substituting the values, we get:
        \[ \text{Percentage} = \left( \frac{2 \text{ feet}}{200 \text{ feet}} \right) \times 100  \]
        - Simplify the fraction:
        \[ \frac{2}{200} = 0.01 \]
        - Multiply by 100 to convert to a percentage:
        \[  0.01 \times 100 = 1\%  \]
        Therefore, the combined length of the remoras is \(\boxed{1}\%\) of the whale's body length.
    };
    
    \node (box3) [boxb, right=of box2] {
        \textbf{GKD-GRPO $\lambda=1$}:\\[0.2cm]
To determine what percentage of the whale's body length the combined length of the remoras is, we need to follow these steps:\par
1. Convert the length of the remoras from inches to feet.\par
2. Add the length of the remoras to the length of the whale.\par
3. Calculate the percentage of the whale's body length that the combined length of the remoras represents.\par
First, we convert the length of the remoras from inches to feet. Since there are 12 inches in a foot, we divide 12 by 12 to get 1 foot per inch. Therefore, the length of each remora is:
\[ 12 \text{ inches} \div 12 = 1 \text{ foot} \]
Since there are 2 remoras, the total length of the remoras is:
\[ 2 \text{ remoras} \times 1 \text{ foot/remora} = 2 \text{ feet} \]
Next, we add the length of the remoras to the length of the whale:
\[ 200 \text{ feet} + 2 \text{ feet} = 202 \text{ feet} \]
Now, we calculate the percentage of the whale's body length that the combined length of the remoras represents. We do this by dividing the total length of the remoras by the length of the whale and then multiplying by 100:\par\[ \left( \frac{2 \text{ feet}}{200 \text{ feet}} \right) \times 100 = 1\% \]
Therefore, the combined length of the remoras is \(\boxed{1}\%\) of the whale's body length.
    };
    \coordinate (midpoint_2_3) at ($(box2.center)!0.5!(box3.center)$);

    \node (box1) [box_hoz, above=of midpoint_2_3,yshift=5.4cm] {
        \textbf{Question}:\\
        Wei saw a 200-foot whale with 2 12-inch remoras attached to it. What percentage of the whale's body length is the combined length of the remoras?
    };
    
    \node (box4) [box_hoz, below=of $(box2.center)!0.5!(box3.center)$,yshift=-5.4cm] {
        \textbf{Judge}:\\
        Response A is more clear and logically structured, breaking down each step with proper mathematical notation and avoiding unnecessary complexity. Response B incorrectly adds the remoras' length to the whale's length, which is irrelevant to the problem and introduces confusion.\par
        Verdict: A wins
    };
    
    \end{tikzpicture}
    \caption{Example of generated answer with Qwen2.5-1.5B after distillation.}
    \label{fig:qwen2_quali2}
\end{figure}

\begin{figure}[htb]
    \centering 
    \begin{tikzpicture}[
        boxa/.style={
            draw,
            rectangle,
            rounded corners,
            fill=green!5,
            text width=0.48\linewidth,
            align=left,
            font=\scriptsize,
        },
        boxb/.style={
            draw,
            rectangle,
            rounded corners,
            fill=red!5,
            minimum height=6.22cm,
            text width=0.48\linewidth,
            align=left,
            font=\scriptsize,
            execute at begin node={\vbox\bgroup\raggedright},
            execute at end node={\vfil\egroup},
            text height=1.9cm,
        },
        box_hoz/.style={
            draw,
            rectangle,
            rounded corners,
            fill=blue!5,
            text width=0.98\linewidth,
            minimum height=1cm,
            align=left,
            font=\scriptsize,
            inner sep=4pt
        },
        node distance=0.1cm
    ]
        
    \node (box2) [boxa] {
        \textbf{Ours}:\\[0.2cm]
        To determine how many fourth-grade boys were at Maple Grove School on Thursday, we need to follow these steps:\par
        1. \textbf{Calculate the number of fourth-grade boys:}\par   
        - The total number of fourth-graders is 154.\par   
        - The number of fourth-grade girls is 80.\par   
        - Therefore, the number of fourth-grade boys is:
        \[ 154 - 80 = 74 \]
        2. \textbf{Determine the number of fourth-grade boys absent:}\par   
        - On Thursday, 6 fourth-grade boys were absent. \par 
        3. \textbf{Calculate the number of fourth-grade boys present:}\par   
        - The total number of fourth-grade boys is 74.\par   
        - The number of fourth-grade boys absent is 6.\par   
        - Therefore, the number of fourth-grade boys present is:\par     
        \[ 74 - 6 = 68 \]\par\par
        Thus, the number of fourth-grade boys at Maple Grove School on Thursday is \(\boxed{68}\).
    };
    
    \node (box3) [boxb, right=of box2] {
        \textbf{GKD}:\\[0.2cm]
To determine how many fourth-grade boys were at Maple Grove School on Thursday, we need to follow these steps:\par \par 1. \textbf{Calculate the number of fourth-grade girls present:}\par    
- Total number of fourth-graders: 154\par    
- Number of fourth-grade girls: 80\par    
- Number of fourth-grade girls present: \( 154 - 3 = 151 \)\par 
2. \textbf{Calculate the number of fourth-grade boys present:}\par    
- Total number of fourth-graders: 154\par    
- Number of fourth-grade boys: \( 154 - 80 = 74 \)\par    
- Number of fourth-grade boys present: \( 74 - 6 = 68 \)\par \par 
Thus, the number of fourth-grade boys present on Thursday is \(\boxed{68}\).
    };
    \coordinate (midpoint_2_3) at ($(box2.center)!0.5!(box3.center)$);

    \node (box1) [box_hoz, above=of midpoint_2_3,yshift=3.1cm] {
        \textbf{Question}:\\
        There are 154 fourth-graders at Maple Grove School. 80 of them are girls. On Thursday, 3 fourth-grade girls and 6 fourth-grade boys were absent. How many fourth grade boys were at Maple Grove School on Thursday?
    };
    
    \node (box4) [box_hoz, below=of $(box2.center)!0.5!(box3.center)$,yshift=-3.1cm] {
        \textbf{Judge}:\\
        Response A provides a clearer and more accurate approach by first determining the total number of boys and then subtracting the absent ones. Response B incorrectly calculates the number of girls present, which could lead to confusion.\par
        Verdict: A wins
    };
    
    \end{tikzpicture}
    \caption{Example of generated answer with Qwen2.5-1.5B after distillation.}
    \label{fig:qwen2_quali3}
\end{figure}

\end{document}